\documentclass[graybox,11pt]{svmult}

\newcommand\norm[1]{\left\| #1 \right\|}

\usepackage{times,epsfig,graphicx}
\usepackage{amsmath,amssymb,amsopn,amsfonts}
\usepackage{algpseudocode}
\usepackage[ruled,noend]{algorithm2e}

\usepackage{esvect}
\usepackage{url}

\makeatletter

\usepackage{natbib}
\renewcommand{\cite}{\citep}

\makeatletter
\DeclareRobustCommand\onedot{\futurelet\@let@token\@onedot}
\def\@onedot{\ifx\@let@token.\else.\null\fi\xspace}

\newcommand{\bfP}{\mathbf{P}}

\newcommand{\bfp}{\mathbf{p}}

\newcommand{\field}[1]{\mathbb{#1}}
\newcommand{\Real}{\field{R}} 
\renewcommand{\I}{\field{I}}
\newcommand{\R}{\Real}

\def\1{{%
    \setbox0\hbox{1}%
    \rlap{\hbox to \wd0{\hss~1\hss}}\box0
}}

\newcommand{\bi}{\begin{itemize}}
\newcommand{\ei}{\end{itemize}}
\newcommand{\bt}[1]{\begin{tabular}{#1}}
\newcommand{\et}{\end{tabular}}
\newcommand{\ba}[1]{\begin{array}{#1}}
\newcommand{\ea}{\end{array}}

\newcommand{\be}{\begin{equation}}
\newcommand{\ee}{\end{equation}}
\newcommand{\bea}{\begin{eqnarray}}
\newcommand{\eea}{\end{eqnarray}}
\newcommand{\ben}{\begin{eqnarray*}}
\newcommand{\een}{\end{eqnarray*}}

\algnewcommand\algorithmicinput{\textbf{Input:}}
\algnewcommand\Input{\item[\algorithmicinput]}
\algnewcommand\algorithmicoutput{\textbf{Output:}}
\algnewcommand\Output{\item[\algorithmicoutput]}

\DeclareMathAlphabet{\mathcal}{OMS}{cmsy}{m}{n}

\begin{document}

\title*{From handcrafted to deep local features}
\institute{Updated version of our preprint}

\author{{\Large Gabriela Csurka, Christopher R. Dance and Martin Humenberger}\\ 
{\large NAVER LABS Europe,  6 chemin de Maupertuis, 38240 Meylan, France} \\ 
{\em \large  firstname.lastname@naverlabs.com} \\
{\em \large www.europe.naverlabs.com}}

\authorrunning{G. Csurka, C. R. Dance and M. Humenberger}
\titlerunning{From handcrafted to deep local features - Preprint}

\maketitle

\abstract{This paper presents an overview of the evolution of local features from handcrafted to deep-learning-based methods, followed by a discussion of several benchmarks and papers evaluating such local features. 
Our investigations are motivated by 3D reconstruction problems, where the precise location of the features is important. As we describe these methods, we highlight and explain the challenges of feature extraction and potential ways to overcome them. We first present handcrafted methods, followed by methods based on classical machine learning and finally we discuss methods based on deep-learning. 
This largely chronologically-ordered presentation will help the reader to fully understand the topic of image and region description in order to make best use of it in modern computer vision applications. In particular, understanding handcrafted methods and their motivation can help to understand modern approaches and how machine learning is used to improve the results. We also provide references to most of the relevant literature and code.}

\section{Introduction}
\label{sec:intro}

In the computer vision literature we encounter local features in two main contexts. In the first context, a local feature is the description of a region of an image in terms of characteristics extracted from that region. Such local features are usually called {\em local descriptors} and are usually represented by vectors of real numbers or binary digits. For example, local descriptors could be obtained by concatenating the pixel intensities or computing a colour histogram of the region.
In the second context, a local feature is a distinctive point or area of an image, and such local features are often called {\em keypoints}, {\em interest points} or {\em anchor points}. It is often important both to find the precise location of such a local feature in the image, which is called {\em detection}, and to extract a corresponding local descriptor.   
Keypoint detection also usually involves the determination of a set of transformations that may be applied to the region in order to make its descriptor invariant to some geometric or photometric transformations. This paper discusses both detection and description. 
For clarification, in the remainder of this paper a {\em local feature} consists of a {\em keypoint/interest point} and its {\em descriptor}.\\

Local features are important for object recognition by the human visual system~\citep{BiedermanPR87Recognition}. They have also been successful in many computer vision applications including face and object detection and recognition, image retrieval, 
motion detection and tracking, depth-map generation, image stitching, camera calibration, 3D reconstruction, 
structure from motion (SfM), visual odometry, and visual simultaneous localisation and mapping (VSLAM). \\

We distinguish three broad categories of local features depending on the applications for which those features were designed~\citep{TuytelaarsFTCGV07Local}. 
First, local features may be designed to have a specific semantic interpretation. For instance edges in aerial images often correspond to roads, while blobs often represent impurities in inspection tasks. 
Second, local features may be designed to be localised accurately and consistently over time. Such local features are important in 
matching (\textit{\textit{e.g.}}~camera calibration and 3D reconstruction) and tracking (\textit{e.g.}~visual odometry) applications. 
Finally, the set of local features extracted from an image can be used as a robust representation for the recognition and retrieval
of objects and scenes. Such local features do not need to have specific semantics or to be accurately localised, as the corresponding applications mainly exploit the statistics of the local-feature set.\\ 

In this paper we are mainly interested in the second category of local features, where the aim is to accurately localise and match keypoints that correspond to the same 3D point viewed in different images.
Ideally, such local features should be robust to variations in \emph{viewpoint} (geometric deformations) and \emph{lighting} (photometric changes).
They should also be robust to image noise, discretisation effects, compression artifacts and blur. 
Furthermore, the descriptors should be \emph{distinctive}.
That is, they should have a rich enough information content that local features corresponding to different 3D points are readily distinguished, while local features corresponding to the same 3D point are readily matched, even in the presence of a large number of such features.
On the other hand, many applications require descriptors that are \emph{compact}.
That is, they should be of low dimension so that they require little memory to store and may be efficiently matched. \\

\begin{figure}[ttt]
\begin{center}
\includegraphics[width=0.9\textwidth]{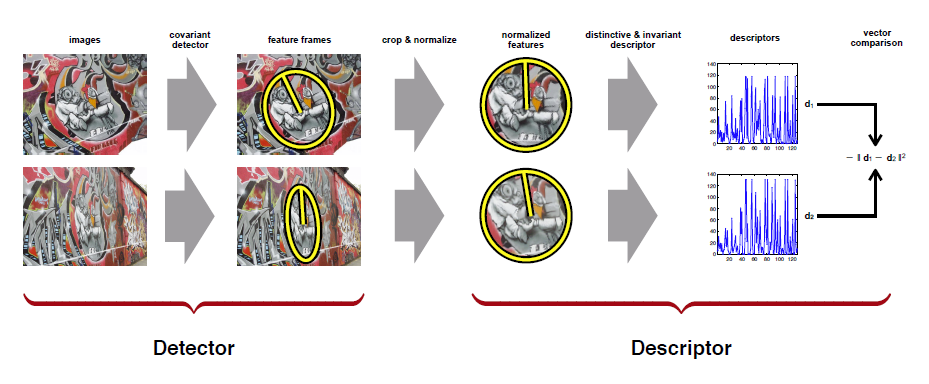}
\caption{Local feature detection and description pipeline. 
A detector is \emph{covariant} to a particular class of geometric transforms if it extracts transformed versions of the same regions when an image is subject to a transform from that class. For instance, regions detected by a scale and rotation covariant detector should scale by $2\times$ and rotate by $30^\circ$ if the image is scaled by $2\times$ and rotated by $30^\circ$. (By courtesy of Andrea Vedaldi.)} 
\label{fig:locfeat}
\end{center}
\end{figure}

Since the geometric relationship between two images of the same scene is usually a projective transformation, one might argue that local features should be invariant to projective transformations, or that they should be invariant to affine transformations, which approximate small perspective transformations well. One might achieve such invariance in the following way, as illustrated in Figure~\ref{fig:locfeat}.
First one defines a canonical view of the patch and estimates the geometric transformation between the image and the canonical view. 
Then one transforms the local region into the canonical view and computes the descriptor on this normalised patch. 
Nevertheless, most local features are only rotation and scale invariant, and they compute normalised patches by first estimating each keypoint's dominant orientation and then its characteristic scale. \\

To address these requirements, a large set of different keypoint detectors 
and descriptors have been proposed in the last three decades. 
In addition to our description, 
the reader should refer to the comprehensive surveys of~\citet{TuytelaarsFTCGV07Local}, \citet{KrigB14Computer} and~\citet{FanB14Local}, as well as the benchmark papers of~\citet{MikolajczykIJCV05Comparison}, \citet{MikolajczykPAMI05Performance}, 
\citet{HeinlyECCV12Comparative}, 
\citet{BalntasCVPR17HPatches} 
and~\citet{SchonbergerCVPR17Comparative}.
Many of these feature detectors and descriptors are integrated in OpenCV\footnote{See the section entitled ``Feature Detection and Description'' in the OpenCV Tutorial at \url{https://docs.opencv.org/3.1.0/d6/d00/tutorial_py_root.html}.} and
 VLFeat\footnote{The Matlab VLFeat library is available at \url{http://www.vlfeat.org/}.}. \\

Aggregates of local features such as the bag of visual words~\citep{SivicICCV03Text,CsurkaSLCV04Visual},
and its extensions such as Fisher Vectors~\citep{PerronninCVPR07Fisher} and VLAD~\citep{JegouCVPR10Aggregating} were the most widely used representations
for image classification and retrieval for a timespan of more than a decade. Therefore a tremendous amount of work on learning, combining,
and improving local features for object recognition, image classification, and retrieval was published. However, detailed presentation of that literature is out of the scope of this paper which instead focuses on local features that may be accurately localised and consistently matched. \\
 
This paper is structured as follows. 
We present work on handcrafted local features (Section~\ref{sec:handcrafted}) 
before discussing methods based on classical machine learning (Section~\ref{sec:learn}).
Then we present advances in detection and description using deep learning (Section~\ref{sec:deeplearn}).
Finally, we summarise the main findings of several benchmark papers (Section \ref{sec:discuss}) and conclude (Section \ref{sec:conclusion}).\\

\section{Handcrafted local features}
\label{sec:handcrafted}

Before the deep-learning revolution, handcrafted local features were a key element of almost all computer vision applications. 
The requirements for these applications were diverse.
Therefore, many different handcrafted detectors and descriptors have been proposed over the last three decades. 
While several of the papers discussed in this section introduce both a detector and a descriptor, it is often possible to pair the detector from one local feature with the descriptor from another. 
For this reason, this section first discusses keypoint detectors (Section~\ref{sec:keypoints}),
then local feature descriptors given by real vectors that we refer to as \emph{real descriptors} (Section~\ref{sec:desc}) 
and finally descriptors given by binary vectors that we refer to as \emph{binary descriptors} (Section~\ref{sec:binary}).
 \\

\subsection{Keypoint detectors}
\label{sec:keypoints}
Handcrafted keypoint detectors may be based on finding corners, analysing intensity derivatives, 
segmentation, mathematical morphology, saliency and normalised intensity edges, as we now describe.\\

The earliest detectors were based on finding corners and on analysing intensity derivatives. Corner detectors find maxima of curvature or abrupt changes in the direction of the tangent to an edge~\citep{RosenfeldB82Digital,WangIVC95RealTime}.
Meanwhile, intensity-derivative-based detectors try to find regions that satisfy certain uniqueness and stability criteria,
and include the Hessian detector~\citep{ZunigaCVPR83Corner} and Harris detector~\citep{HarrisAnvey88Combined} as notable examples.
The Hessian detector, also referred to as the determinant of Hessian (DoH) method,
is used in the popular speeded up robust features (SURF) algorithm~\citep{BayECCV06Surf}, where for efficiency, the Hessian is roughly approximated with a set of box filters and no smoothing is applied when going from one scale to the next.
Both Harris and Hessian methods were extended by \citet{MikolajczykIJCV04Scale} to handle 
affine invariance. 
Furthermore, one of the most popular keypoint detection methods, the scale-invariant feature transform (SIFT) detector~\citep{LoweIJCV04Distinctive}, can be seen as an intensity-derivative-based method. In particular, the SIFT detector uses the difference of Gaussians (DoG) to detect local extrema and the DoG can be viewed as an approximation of the Laplacian of a Gaussian. \\

Segmentation techniques have also been employed for detection. Such methods either work with junctions on the boundaries of homogeneous regions~\citep{LiuPR90Moment} or they work with the homogeneous regions themselves~\citep{CorsoCVPR05Coherent}. 
Similarly, the maximally stable extremal regions (MSER) detector~\citep{MatasBMVC02Robust}, which was developed for estimating disparities in wide-baseline stereo, uses watershed-like segmentation to extract homogeneous intensity regions which are stable
over a wide range of thresholds. \\

Several detectors are based on ideas from mathematical morphology. For instance, SUSAN (univalue segment assimilating nucleus)~\citep{SmithIJCV97SUSAN} computes the fraction of pixels within a neighbourhood which have similar intensity to the center pixel. Corners can then be localised by applying a threshold to this measure and selecting local minima. FAST (features from accelerated segment test)~\citep{RostenECCV06Machine} is an extension of SUSAN, whose keypoint detector is significantly faster.
The method relies on a set of pixels in a circular pattern to determine a keypoint, and makes comparisons between the intensities of pixels on the circle and the intensity of the pixel at the center of the circle. If a number of consecutive pixels around the circle are consistently brighter than the centre or consistently darker than the centre, then the central pixel is considered to be a good candidate. The process concludes with non-maximum suppression. As originally proposed, FAST is not a scale-space detector, so \citet{LepetitPAMI06Keypoint} extended it to perform scale selection with the Laplacian function. \\

Salient-region-based detectors exploit the notion that keypoints should exhibit local attributes that are unpredictable
compared to the surrounding region. For instance, \citet{KadirECCV04Affine} proposed to measure the change in entropy of a grey-value histogram computed in a set of neighbourhoods of variety of positions, scales and affine shapes. Wavelet transformations have also been considered for multi-resolution keypoint detection \citep{SebeIVC03Evaluation}.\\

Rather than working directly with image intensities, EdgeFoci\footnote{The EdgeFoci page can be found at \url{http://research.microsoft.com/en-us/um/people/larryz/edgefoci/edge_foci.htm}.}~\citep{ZitnickICCV11Edge} works with normalised intensity edges, as its authors hypothesize that such edges are more robust to nonlinear lighting variations and background clutter. In particular, they define an \emph{edge focus} as a point that is equidistant from two edges whose edge orientations are perpendicular, and use such edge foci as keypoints. \\

\subsection{Real descriptors}
\label{sec:desc}

\begin{figure}[ttt]
\begin{center}
\includegraphics[width=0.85\textwidth]{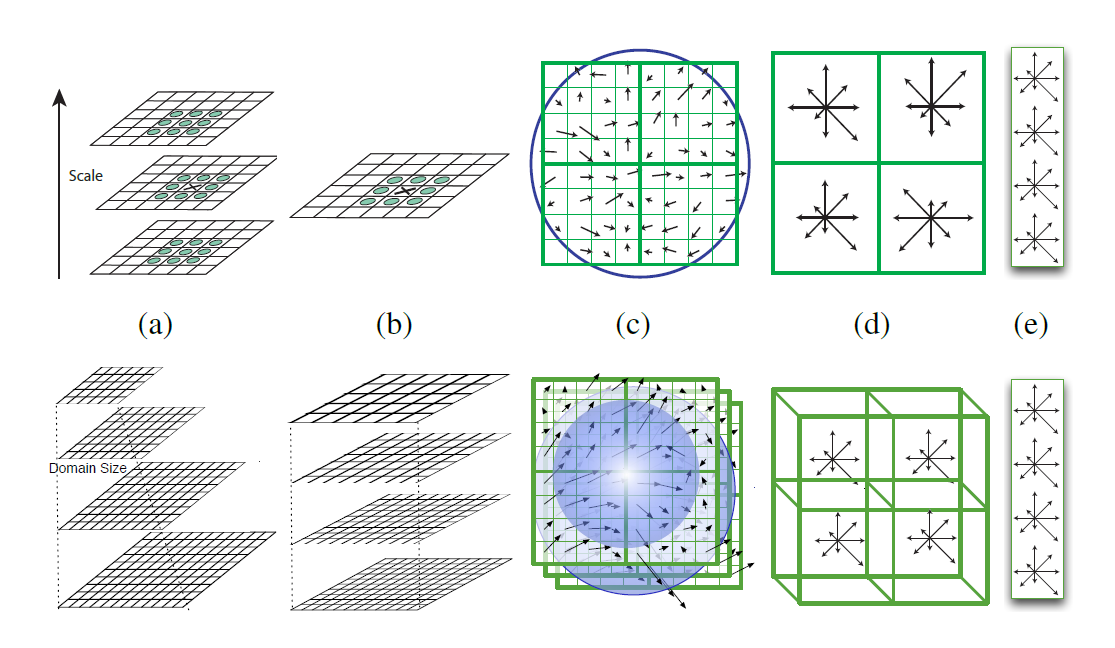}
\caption{Top Row: The SIFT~\citep{LoweIJCV04Distinctive} detection and description process. 
(a) The difference of Gaussians is computed at multiple scales; 
(b) A scale is selected for each keypoint; 
(c) Gradient orientations are computed at that scale; 
(d) The orientations are spatially pooled;
(e) This yields histograms that are concatenated and normalised to form the descriptor. 
Bottom Row: The DSP-SIFT~\citep{DongCVPR15DomainSize} detection and description process.  
(a,b) Patches of different sizes are re-scaled;
(c,d) Gradient orientations are computed and pooled across locations and scales;
(e) These are concatenated yielding a descriptor of the same dimension as
the ordinary SIFT descriptor.
(By courtesy of Jingming Dong.)}
\label{fig:DSPSIFT}
\end{center}
\end{figure}

One of the most widely used local feature descriptors is SIFT~\citep{LoweIJCV04Distinctive,OteroIPOL14Anatomy}. 
In the literature, SIFT often refers to the whole pipeline shown in Figure~\ref{fig:DSPSIFT} (top row), including the detector
as well as the descriptor which is based on local gradient histograms computed on a $4\times 4$ grid. 
SIFT has been extended in several ways.
These include the gradient location and orientation histogram (GLOH)~\citep{MikolajczykPAMI05Performance}, which considers a log-polar grid on which the gradients are averaged to compute the histograms, coloured SIFT (CSIFT)~\citep{AbdelHakimCVPR06More} which exploits colour-invariant characteristics, 
domain-size pooling SIFT (DSP-SIFT)~\citep{DongCVPR15DomainSize}, which pools SIFT descriptors across scales, and 
scale-less SIFT (SLS)~\citep{HassnerPAMI17Sifting}, which is a subspace representation of SIFT across multiple scales. \\

Whereas SIFT uses a scale space obtained by smoothing and downsampling the input image, the idea underlying DSP-SIFT~\citep{DongCVPR15DomainSize} is to instead use a \emph{size-space} which is obtained by maintaining the same scale as the input image, but considering subsets of it of variable size. 
Furthermore, while SIFT descriptors are constructed at
a selected scale and gradient orientations are pooled in its spatial neighbourhood, DSP-SIFT considers patches of different sizes that are re-scaled and gradient orientations are pooled across locations \emph{and} scales (see Figure~\ref{fig:DSPSIFT}, bottom row). Note that domain-size pooling (DSP) can also be applied to features other than SIFT. \\

SLS~\citep{HassnerPAMI17Sifting} represents each pixel with a set of SIFT descriptors extracted at multiple scales. The authors show that SLS gives far better matches than descriptors computed at a single selected scale. As this improvement comes at a significant computational cost, the authors propose to represent each set of SIFT descriptors by a low-dimensional, linear subspace and a subspace-to-point mapping is used to get the final descriptors.\\
 
Several papers propose descriptors that are faster than SIFT yet still allow for reliable matching. 
Notably, SURF~\citep{BayECCV06Surf} computes descriptors using Haar filter responses obtained from integral images. Also,  
KAZE (which is the Japanese word for \textit{wind})~\citep{AlcantarillaECCV12OKaze} uses a similar pipeline to SURF, except that it works in a nonlinear scale space. This nonlinear scale space is built using efficient additive operator splitting techniques and variable conductance diffusion. 
Accelerated KAZE\footnote{(A)KAZE is available at \url{http://www.robesafe.com/personal/pablo.alcantarilla/kaze.html}.}~\citep{AlcantarillaBMVC13Fast} uses fast explicit diffusion embedded in a pyramidal framework to dramatically speed up detection. \\

\subsection{Binary descriptors}
\label{sec:binary}

With the spread  of mobile and embedded vision systems, the demand for efficient detection and matching of image features
grew. Also, in mobile applications, it is desirable to limit the amount of data sent over the network in order to keep latency and
costs down.  Moreover, in applications such as tracking and visual simultaneous localisation and mapping (VSLAM), keypoint detection and description often have to be done in real time. 
These facts have motivated many authors to propose binary descriptors, which require less storage than real descriptors and can be efficiently matched using Hamming distance. \\

\begin{figure}[ttt]
\begin{center}
\includegraphics[width=\textwidth]{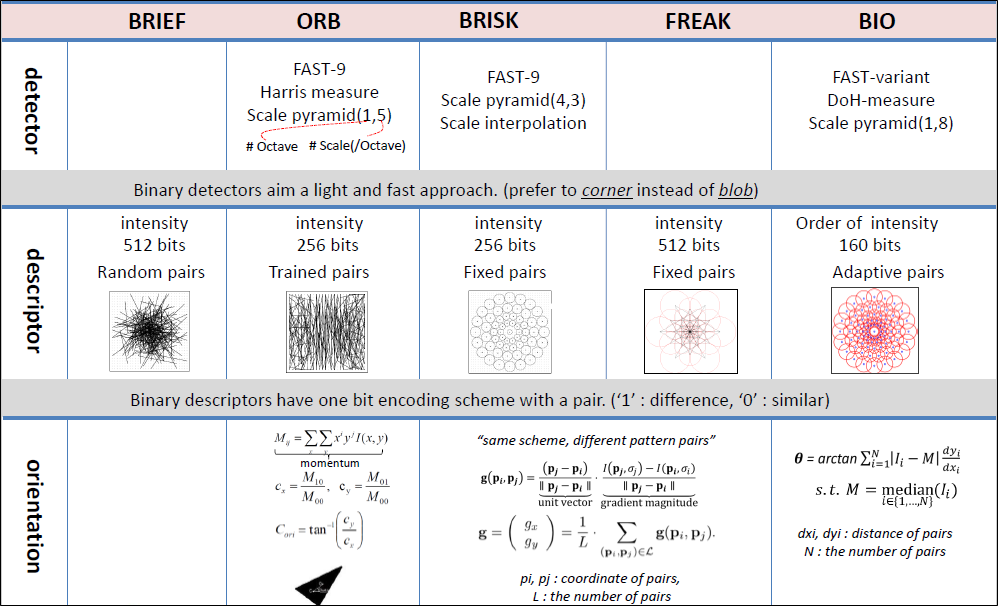}
\caption{An illustrative summary of some of the popular binary descriptors from~\citet{ChoiURAI15Approach}. 
(By courtesy of Yukyung Choi.)} 
\label{fig:binary}
\end{center}
\end{figure}

One way to build binary descriptors is to binarise existing real descriptors through quantisation~\citep{GongCVPR11Iterative} or hashing~\citep{BrownCVPR05MultiImage,GionisVLDB99Similarity,ShakhnarovichICCV03Fast,WeissNIPS08Spectral,KulisICCV09Kernelized,StrechaPAMI12LDAHash}. 
Alternatively, binary descriptors may be extracted directly from image patches.
Such binary descriptors include LBP, the census transform, BRIEF, ORB, BRISK, FREAK and BIO, as we now discuss (see also Figure~\ref{fig:binary}). \\

Local binary patterns (LBP)~\citep{OjalaPR96Comparative,PietikainenB11Computer} 
and the census transform~\citep{ZabihECCV94Nonparametric} both create a descriptor by comparing the intensity of a given pixel with each of its neighbours' intensities, and encoding 1 if the value is greater and 0 otherwise. 
Despite their simplicity, LBP and the census transform have proved quite powerful, so they have inspired a large set of variants. \\
  
The binary robust independent elementary feature (BRIEF) descriptor~\citep{CalonderECCV10Brief} uses a randomly-selected distribution of   
point-pairs relative to a central point to create the descriptor. Oriented BRIEF (ORB) features~\citep{RubleeICCV11Orb}, add rotational invariance to BRIEF by determining corner orientation using FAST~\citep{RostenECCV06Machine}. \\

Binary robust invariant scalable keypoints (BRISK)~\citep{LeuteneggerICCV11Brisk}
is another popular descriptor that uses a sampling pattern consisting of disks of variable sizes arranged
in four concentric rings. The pattern is pre-rotated to a characteristic direction to achieve rotation invariance.
Finally, the oriented sampling pattern is used to obtain pairwise brightness comparisons
which are assembled into the binary descriptor.\\

The fast retina keypoints (FREAK) descriptor\footnote{Code available at \url{https://github.com/kikohs/freak}.}~\citep{AlahiCVPR12Freak} uses a 
multi-resolution pixel-pair sampling pattern with trained pixel-pairs.
This design mimics the human eye in the sense that it has high resolution in the fovea 
and lower resolution in the periphery. \\

The robust binary feature using intensity order (BIO) descriptor~\citep{ChoiACCV14Robust} was inspired by the local intensity order pattern (LIOP) descriptor\footnote{The code for LIOP is available at \url{http://zhwang.me/publication/liop/index.html}.}~\citep{WangICCV11Local,WangPAMI16Exploring} which encodes local intensity ordering information. The authors use a FAST-like binary comparison test and detect keypoints using a fast approximation of the determinant of Hessian (DoH). As ordinal descriptors are insensitive to moderate rank-order errors, they can be quantised into binary descriptors without
noticeably degrading performance. \\

 \begin{table}[t]
 \caption{Popular datasets for local feature detector and descriptor training and evaluation.}
 \label{tab:patchDataset}
\begin{center}
\bt{|lcccl|} 
\hline
Dataset   & \,\,\,  & Nature of ground truth & \,\,\,  & Reference  \\
\hline
Oxford-Affine & & homographies & & \citep{MikolajczykPAMI05Performance}  \\
Photo-Tourism & & corresponding patches & & \citep{WindercCVPR07Learning} \\
Fountain and Herzjesu & & visibility and optical flow & & \citep{StrechaCVPR08Benchmarking}\\
EdgeFoci & & image sequences & & \citep{ZitnickICCV11Edge} \\
Cornell BigSfM & & 3D points, tracks, camera info & & \citep{CrandallPAMI13SfM} \\
Hannover & & image sequences, homographies & & \citep{CordesCAIP13HighResolution} \\
RomePatches & &  corresponding patches, image labels & &  \citep{LiECCV14CLocation,PaulinICCV15Local} \\
1DSfM & & 3D points  and camera info & & \citep{WilsonECCV14Robust} \\
DaLI & & object deformations & & \citep{SimoSerraIJCV15Dali} \\
WebCam & & image sequences & & \citep{VerdieCVPR15TILDE} \\
HPatches & &  corresponding patches,  homographies & & \citep{BalntasCVPR17HPatches} \\
HSequences & & image pairs, homographies & & \citep{LencECCV18Large} \\
\hline
\et
\end{center}
\end{table}

\section{Local features based on classical machine learning}
\label{sec:learn}
We now discuss local feature detectors and descriptors based on classical machine learning, as opposed to \emph{deep} learning
which is discussed in Section~\ref{sec:deeplearn}. 
As with machine learning methods in general~\cite{MohriB12Foundations}, such methods involve training on a given dataset in the expectation that this will lead to good performance on new data drawn from a similar distribution. 
The training procedure can lie anywhere on a spectrum from unsupervised to supervised.
On the one hand, purely unsupervised methods need no labelled data. 
When such unsupervised methods are applied to learning local descriptors, the main idea is often to adapt handcrafted features to the given dataset, for instance by projecting them into a well-chosen low-dimensional space. 
On the other hand, supervised learning methods require labelled data. 
In this context, positive labels usually corresponds to ``matched keypoints'', which are pairs of patches representing different views of the same 3D point. 
Such matched keypoints can easily be generated synthetically, or extracted from sequences of images of a given scene by leveraging geometric consistency (\textit{i.e.} a known homography or an essential matrix relating different views).
Table~\ref{tab:patchDataset} lists the most popular datasets used to train or evaluate local features,
which include Oxford-Affine\footnote{Oxford-Affine is available at \url{http://www.robots.ox.ac.uk/~vgg/data/data-aff.html}.}, 
Photo-Tourism\footnote{Photo-Tourism dataset page \url{http://matthewalunbrown.com/patchdata/patchdata.html}.}, 
Fountain and Herzjesu\footnote{Fountain and Herzjesu can be downloaded from \url{https://cvlab.epfl.ch/data/keypoint}.}, 
Cornell BigSfM\footnote{Cornell BigSfM is available at \url{http://www.cs.cornell.edu/projects/bigsfm/}.}, 
1DSfM\footnote{1DSfM dataset page \url{http://www.cs.cornell.edu/projects/1dsfm/}.}, 
RomePatches\footnote{RomePatches is available at \url{http://lear.inrialpes.fr/people/paulin/projects/RomePatches/}.} 
and HPatches\footnote{HPatches data and benchmark on \url{https://github.com/hpatches}.}.

As in the previous section, we first discuss detectors (Section~\ref{sec:learndet}), before moving on to real and binary descriptors (Sections~\ref{sec:learndesc} and~\ref{sec:learn_binary}).

\subsection{Learning detectors}
\label{sec:learndet}
Several papers have considered using classical machine learning to \emph{speed up} detection while finding the same keypoints as handcrafted methods~\citep{RostenECCV06Machine,LeuteneggerICCV11Brisk,RubleeICCV11Orb}.
Also, \citet{HartmannCVPR14Predicting} learned a classifier that predicts which keypoints are likely to be discarded when matching descriptors among those extracted by a standard keypoint detection algorithm, namely DoG. By using a random forest to learn such ``matchability'', they showed that the approach can considerably improve and speed up the feature matching stage of a SfM pipeline. However this approach remains limited by the quality of the initial keypoint detector.\\

Other papers have focused on using learning to improve detector \emph{repeatability}. Given a collection of keypoints detected by a standard detector, \citet{StrechaDAGM09Training} demonstrate higher repeatability by using a WaldBoost classifier to keep only keypoints that are known to be useful for the task at hand. For instance, in the task of image matching in an urban environment, their classifier learns to focus on more-stable man-made structures and to ignore objects that undergo regular changes such as vegetation and clouds. \\

Meanwhile~\citet{VerdieCVPR15TILDE} proposed the temporally invariant learned detector (TILDE), which was designed for repeatable keypoint detection in the presence of drastic illumination changes caused by variations in weather, season and time-of-day, to which keypoint detectors tend to be highly sensitive. 
To achieve this goal, the authors worked with a collection of sequences of webcam images, where each sequence consists of images acquired at the same location but different times.
Using this collection as training data, they learned a variety of regressors to predict how likely a SIFT keypoint detected in one image from a given sequence will also be detected at a nearby point in another image from that sequence. 
They compared regressors based on piecewise-linear functions, the LeNet-5 CNN~\citep{LecunPIEEE98Gradient} and boosted regression trees.
While SIFT keypoints were used during training, at runtime the regressor was passed over the entire image, giving a ``score map''.
Then points with a score are selected and subjected to non-maximum suppression. 
The results showed that using piecewise-linear functions as a regressor gave consistently more reliable keypoints than alternative regressors and than known keypoint detectors such as SURF and MSER. As discussed in Section~\ref{sec:discuss}, TILDE remains a state-of-the-art approach to detection in the presence of illumination changes, however it is limited to situations where only keypoints with a common scale are matched.
\\

\subsection{Learning real descriptors}
\label{sec:learndesc}
\begin{figure}[ttt]
\begin{center}
\includegraphics[width=0.95\textwidth]{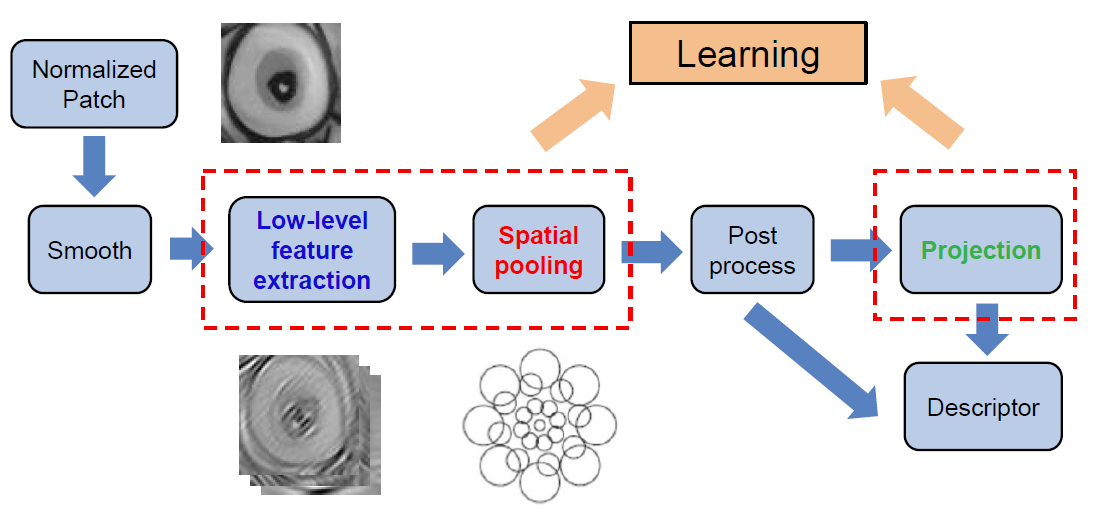}
\caption{Discriminative descriptor learning by optimising  
spatial pooling and feature embedding in \citet{BrownPAMI10Discriminative}. 
(By courtesy of Bin Fan.)} 
\label{fig:Brown}
\end{center}
\end{figure}

Early descriptor-learning methods included unsupervised PCA-SIFT~\citep{KeCVPR04More}, which uses principal component analysis (PCA) to embed the gradient image of a patch into a new space, and supervised methods using randomised trees~\citep{LepetitPAMI06Keypoint} and boosting~\citep{BabenkoICCV07Task} to learn feature representations from matching and non-matching local patch pairs.
More recently, \citet{BrownPAMI10Discriminative} proposed a model which builds on top of handcrafted low-level features, such as steerable filter banks or gradient orientation maps with different spatial pooling, as in the SIFT and DAISY descriptors. 
Considering both linear and nonlinear transforms for dimensionality reduction, they used linear discriminant analysis (LDA) to select the pooling parameters and to obtain low-dimensional representations (see Figure~\ref{fig:Brown}).
Meanwhile, \citet{PhilbinECCV10Descriptor} learned both linear and nonlinear discriminative projections into lower dimensional spaces
using a margin-based cost function, which aims to separate matching descriptors from non-matching descriptors. Training
data was generated automatically by leveraging geometric consistency. \\

More recently, \citet{SimonyanPAMI14Learning} proposed convex large-margin formulations of two tasks in descriptor design:
the selection of spatial pooling regions, for which the authors use $L_1$-regularisation;
and descriptor dimensionality reduction, for which they use nuclear-norm regularisation. 
Their method, ConvOpt\footnote{Code available for ConvOpt on \url{http://www.robots.ox.ac.uk/~vgg/research/learn_desc/}.}, 
applies a stochastic optimisation technique called regularized dual averaging~\citep{XiaoJMLR10Dual} that is well suited to non-smooth sparsity-inducing cost functions. 
The authors further proposed a weakly-supervised extension, where unlabelled data is exploited using an optimisation objective inspired by the large margin nearest neighbour (LMNN) approach~\citep{WeinbergerJMLR09Distance}. \\

\citet{WangECCV14Affine} aimed to make descriptors robust to affine distortions introduced by viewpoint changes. 
Their method performs PCA on a collection of affine-warped versions of an input patch. 
A projection matrix is computed from the first few principal components, in order to describe
this collection of affine-warped patches as a linear subspace. 
To convert this matrix into a descriptor vector, they employ a simple \emph{subspace-to-point mapping}
which consists in stacking the upper-triangular elements of the projection matrix into a vector, after scaling the diagonal
entries by $1/\sqrt{2}$.\\

Only a few papers have attempted to directly design descriptors to be robust to deformations that are more general than affine transformations. 
In the case of images of objects that have undergone nonrigid deformations, there have been two main approaches to finding correspondences given descriptors that are not robust by design.
One approach is to enforce global consistency \citep{ChengCVPR08Deformable,SanchezRieraCVPR10Simultaneous,TorresaniECCV08Feature}
and the other is to include segmentation information in the descriptor and then solve complex optimisation problems to establish matches~\citep{TrullsCVPR13Dense}.   
Arguably, a better option is to build deformation-invariant descriptors. 
As such, DaLI\footnote{Code for DaLI is available at \url{http://www.iri.upc.edu/people/esimo/research/dali/}.} 
(deformation and light invariant) \citep{SimoSerraIJCV15Dali} uses methods from \emph{diffusion geometry} \citep{GebalCGF09Shape,SunCGF09Concise} to build a descriptor for 2D image patches which is invariant to nonrigid deformations and photometric changes. A patch is described in terms of the heat it dissipates onto its neighbourhood over time. 
To ensure compact descriptors, DaLI uses PCA for dimensionality reduction. 
Experimental results demonstrate that the DaLI descriptor can simultaneously handle quite complex photometric changes and spatial warps.\\

\subsection{Learning binary descriptors}
\label{sec:learn_binary}
There is a broad literature on learning binary descriptors for specific applications, such as face recognition~\citep{LeiPAMI14Learning,LuPAMI15Learning,LuPAMI18Simultaneous,DuanPAMI17ContextAware}.
However, this review focusses only on methods that may be relevant for SfM and 3D reconstruction, which include LDAHash, D-BRIEF, RI-LBD, BinBoost, BOLD and RMGD, as we now discuss.\\

LDAHash \citep{StrechaPAMI12LDAHash} computes a projection matrix for a given source of handcrafted descriptors, such as SIFT or SURF, using linear discriminant analysis (LDA). LDA chooses this projection to minimize the ratio of intra-class variance to inter-class variance. 
Finally, the projections are optimally thresholded to give binary vectors.  
In D-BRIEF\footnote{The code for D-BRIEF is available at \url{https://cvlab.epfl.ch/research/detect/dbrief}.}~\citep{TrzcinskiECCV12Efficient} the training data is used to learn linear projections that map image patches to a more discriminative subspace.
In order to obtain binary descriptors, the projected patches are simply thresholded.  
More recently, \citet{DuanTIP17Learning} proposed the rotation-invariant local binary descriptors (RI-LBDs), 
in which each local patch is first categorized into a ``rotational binary pattern''. The orientation for each such pattern and a projection matrix that maps each image patch into a binary code are learned jointly.   \\

BBoost\footnote{Source code for BBoost is available at \url{https://cvlab.epfl.ch/research/detect/binboost}.}
features~\citep{TrzcinskiPAMI15Learning} are low-dimensional but highly discriminative
descriptors computed with a boosted binary hash function. They use weak learners which pool image gradients over particular regions, inspired by handcrafted descriptors like BRIEF~\citep{CalonderECCV10Brief}. 
Similarly, the ring-based multi-grouped descriptor (RMGD)~\citep{GaoTIP15Local} uses pooling over a polar grid, and it performs boosting to select from the set of all binary comparisons between pairs of grid cells.
A circular integral image is used for fast calculation of the binary descriptor. 
To increase discriminativeness and robustness, the RMGD is built with multiple image properties including intensity, $x$- and $y$-gradients,
gradient magnitudes, orientations and soft-assigned gradient orientations.
The receptive fields descriptor (RFD)~\citep{FanTIP14LReceptive} uses gradient-orientation maps summed over two kinds of receptive fields, namely rectangular pooling regions or Gaussian pooling regions. Instead of selecting these regions by boosting, RFD selects them by a greedy approach, according to their distinctiveness and correlations. \\

Whereas most binary descriptors are constructing using the \emph{same} set of measurements for every input patch, binary online learned descriptors (BOLD)\footnote{Open source implementation of BOLD is available at \url{http://vbalnt.io/projects/bold/}.}~\citep{BalntasCVPR15Bold,BalntasPAMI18Binary} \emph{adapt} the set of measurements depending on the input patch. 
This adaptation is motivated by the observation that some of the pairwise intensity comparisons made by conventional binary descriptors are unstable to small affine perturbations, for some patches. 
The adaptation is accomplished by synthesizing multiple small random perturbations of the given input patch.
Inspired by LDA, the authors treat these perturbed versions as coming from a single ``class'', 
and select comparisons leading to small intra-class distances but large inter-class distances. 
The selection of comparisons leading to large intra-class distances is made offline from a large set of binary tests using random patches.  
This online descriptor adaptation process can also be applied to other binary descriptors, and the authors demonstrate that it leads to a consistent improvement in precision-recall curves when applied to BRIEF, ORB and BBoost descriptors. \\

\section{Deep-learning-based local features}
\label{sec:deeplearn}


\emph{Deep learning} is an approach to machine learning that involves mapping an input through a cascade of nonlinear processing layers to produce an output~\citep{DengFTSP14Deep,SchmidhuberNN15Deep,LeCun15Deep,GoodfellowB16Deep}. The cascade is said to be ``deep'' if it has many ($\ge 3$) layers. By automatically learning features across multiple layers, such a system can learn complex functions mapping raw data to outputs directly, without having to rely on handcrafted features.
Often, the layers are those of an artificial neural network, trained by some variant of gradient descent, in which gradients are computed by backpropagation. 
In the field of computer vision, this network is often a convolutional neural network (CNN) with an image as input.
Such a CNN consists of a succession of convolutional layers, whose units (neurons) have learnable weights, and other intermediate layers which play a variety of functions, such as introducing nonlinearities, pooling to downsample and normalising activations across a batch of inputs.\\

In the past six years, deep learning has revolutionized computer vision, enabling huge improvements in the state-of-the-art for tasks such as image recognition~\citep{KrizhevskyNIPS12Imagenet,HeCVPR16Deep,HuangCVPR17Densely} and pushing the community to propose deep-learning methods for most tasks associated with local features. 
As in the previous two sections, our discussion of such methods begins by considering keypoint detection (Section~\ref{sec:deepkp}),
real descriptors (Section~\ref{sec:deepdesc}) and binary descriptors (Section~\ref{sec:deepbin}).
However, we conclude the section with a discussion of methods for end-to-end detection and description (Section~\ref{sec:deep}).

\subsection{Learning detectors with CNNs}
\label{sec:deepkp}

\citet{LencX16Learning} discuss covariant point detectors\footnote{Matlab code from the work of~\citet{LencX16Learning} is available at \url{https://github.com/lenck/ddet}.}. 
The authors treat detection as a regression problem in which one learns a function $\phi : \mathcal{X} \rightarrow\mathcal{G}$ that maps image patches from a set $\mathcal{X}$ to transformations from a group $\mathcal{G}$. They use a loss that encourages the function $\phi$ to approximately satisfy the \emph{covariance constraint}, which requires that 
\begin{align*}
\phi(gx) = g\phi(x) \quad \text{for all image patches $x\in\mathcal{X}$ and transformations $g \in \mathcal{G}$.}
\end{align*}
Approximating such functions $\phi$ with CNNs resembling the compact LeNet model of~\citet{LecunPIEEE98Gradient}, the authors learn three detectors.
The first two detectors are covariant with the group of translations, \text{i.e.} they are corner detectors. The third detector is covariant with the group of translations and 2D rotations. The authors call the latter detector \textsc{RotNet}, but we shall use the terminology of~\citet{ZhangCVPR17Learning} and call it CovDet. 
To use the CovDet detector, the authors apply the CNN $\phi$ to the full image, and each pixel votes for a single rotation and translation. 
These votes are accumulated and used as confidence scores.
Only rotations and translations whose confidence score exceeds a threshold are subjected to non-maximum suppression and retained as detected keypoints.
Their results show that CovDet performs noticeably better than SIFT when recovering the relative orientation of randomly rotated patch pairs.\\

\begin{figure}[ttt]
\begin{center}
\includegraphics[width=\textwidth]{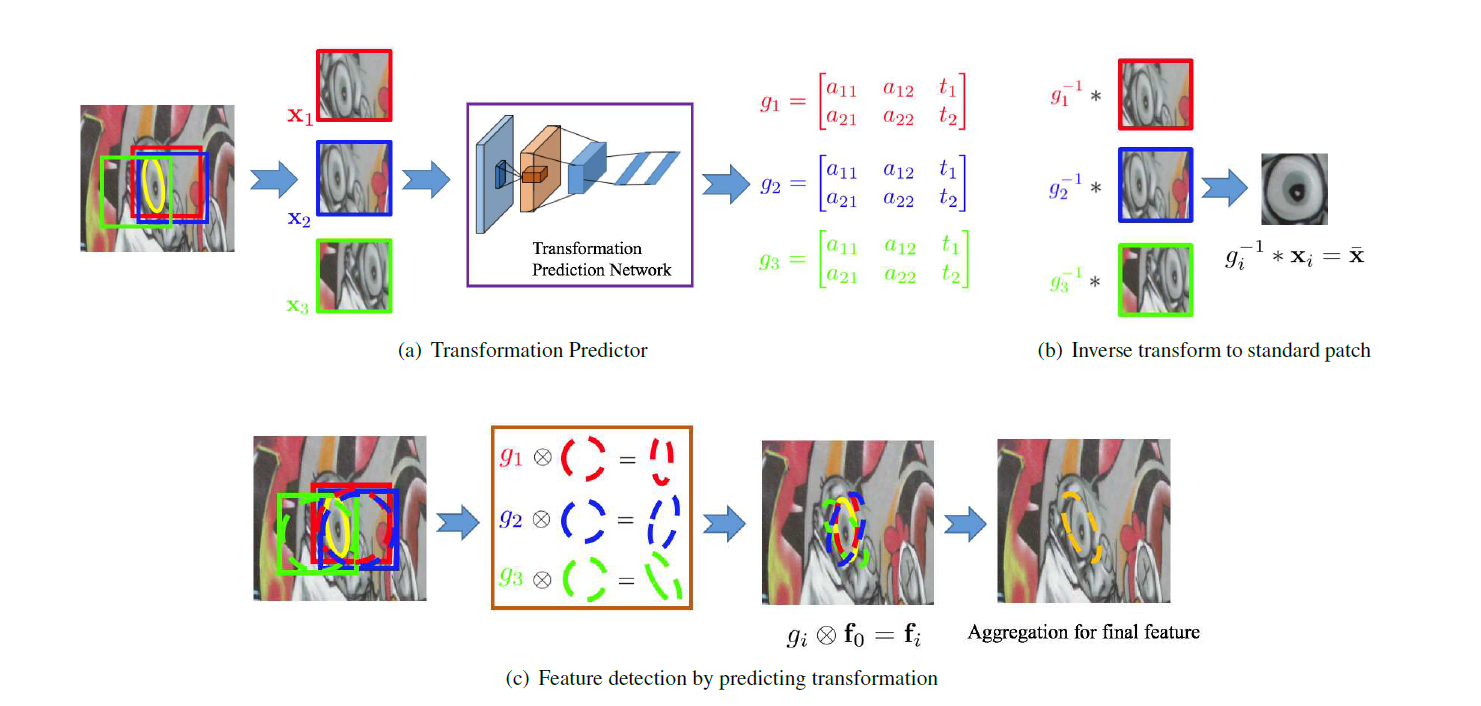}
\caption{Illustrating the TCovDet~\citep{ZhangCVPR17Learning} keypoint detection framework. (a) The transformation prediction network predicts the geometric transformation of the patch. (b) The inverse of the predicted transformation is such that it warps the patch to a ``{\em standard patch}''. (c) The predicted transformations are aggregated to vote for the locations and shapes of keypoints. (By courtesy of Xu Zhang.)} 
\label{fig:TCovDet}
\end{center}
\end{figure}

The transformation covariant local feature detector (TCovDet)\footnote{TensorFlow code for TCovDet is available at \url{https://github.com/ColumbiaDVMM/Transform_Covariant_Detector}.}~\citep{ZhangCVPR17Learning} is an extension of CovDet.
Like CovDet, TCovDet treats detection as a regression problem, learning a CNN $\phi$, called the ``transformation prediction network'' or ``transformation regressor'', that maps input patches to transformation matrices and selecting keypoints based on voting for transformations. 
However, the network in TCovDet is trained using patches selected by TILDE, which the authors call ``standard patches''.
Also, the training objective is to recover 24 randomly-selected affine transforms per standard patch, rather than just translations and rotations as in CovDet.
Furthermore, the loss is augmented by requiring that the original ``untransformed'' patch detected by TILDE maps to the identity transformation. The authors argue that this augmented loss resolves a major drawback with CovDet, which is that the regressor $\phi$ that CovDet learns may not be unique.
The results show that TCovDet improves repeatability scores relative to 13 other detectors on three large datasets. 
Furthermore, TCovDet improves matching scores against four of those detectors in two out of the three datasets.
\\

\citet{SavinovICCV17Quad} treated detector design as the problem of learning a response function, which maps patches to real numbers that determine a \emph{ranking} of image points. The keypoints detected are those points at which the value of the response function is in the top or bottom quantiles of the set of all responses after non-maximum suppression. Such a detector is considered to be good if these quantiles are preserved under the transformations to which the detector is intended to be invariant. The authors propose a variety of neural networks implementing such response functions that they collectively call \emph{quad-networks} since these networks are trained using a quadruplet loss. A \emph{quadruplet loss} is a loss which takes the network outputs for four patches as input. Specifically, say we are given a set of four patches $\mathcal{P}:=\{p,q,p',q'\}$ consisting of a pair of patches $p,q$ corresponding to distinct world points and the same pair of patches $p',q'$ after a transformation that we want the detector to be invariant to. The aim is to train weights $w$ of the response function $H_w(\cdot)$ to minimise the loss 
\begin{align*}
\text{loss}_w(\mathcal{P}) := \max\{0,1-R_w(\mathcal{P})\} \quad \text{where $R_w(\mathcal{P}) := (H_w(p)-H_w(q)) (H_w(p')-H_w(q')).$}
\end{align*}
This loss vanishes if $R_w(\mathcal{P})\ge 1$ and is positive otherwise. In order for $R_w(\mathcal{P})$ to be large, it is necessary that either patches $p,p'$ both have much larger responses than patches $q,q'$ respectively, or patches $q,q'$ both have much higher responses than patches $p,p'$ respectively.\\

The authors compared detectors based on a variety of network architectures with the DoG detector, but with no other handcrafted or learned detector, in two settings. The first setting involved learning a detector to match RGB images with RGB images. In this setting the authors consider a linear network and a convolutional network with a single hidden layer, finding that both networks outperform the DoG detector with except in one test. The exceptional test measured the robustness of matching to JPEG compression artefacts, which was a transformation not included in training data. The second setting involved matching RGB images with depth images. In this setting they considered three networks: a deep network with 10 convolutional layers interleaved with exponential linear units (ELU) and batch normalisation layers; a shallow fully-connected network; and a deep fully-connected network. The deep fully-connected network outperformed the other detectors on this task when the number of keypoints to be detected per image was moderate (between 500 and 1500).


\subsection{Learning real descriptors with CNNs}
\label{sec:deepdesc}
We now discuss the use of CNNs to learn real descriptors targeting applications involving matching pairs of patches corresponding to the same 3D point. This topic has attracted at least 12 research papers to date. In fact, the earliest work on this topic~\citep{JahrerCVWW08Learned} predates AlexNet~\citep{KrizhevskyNIPS12Imagenet}, the network whose performance on ImageNet was a key driver of the boom in popularity of CNNs in computer vision. We begin with papers exploring the use of intermediate activations of AlexNet as descriptors, before discussing approaches based on metric learning, in which the network learns not only descriptors but also a function giving a distance between descriptors. The remainder of the section considers CNNs whose outputs are descriptors to be compared with Euclidean distance (with one exception). We structure the discussion around the loss used to train these CNNs, considering pairwise losses, triplet losses, global losses and finally histogram losses.\\


\textbf{AlexNet Activations.} 
The success of AlexNet~\citep{KrizhevskyNIPS12Imagenet} inspired many authors to use the activations of its intermediate layers as descriptors for other datasets and tasks other than the ImageNet classification task for which it was designed. 
\citet{DonahueICML14Decaf} showed that such descriptors gave results that surpassed the state of the art at the time on tasks including domain adaptation, fine-grained recognition and scene-type recognition (with \emph{abbey}, \emph{dinner}, \emph{mosque} and \emph{stadium} as categories). 
Meanwhile, \citet{LongNIPS14Convnets} successfully applied the intermediate activations of a network almost identical to AlexNet to 
the tasks of intra-class alignment, keypoint prediction and keypoint classification. \\

\citet{FischerX14Descriptor} were the first to use the intermediate activations of AlexNet as descriptors for the task of matching patches corresponding to the same 3D point. 
Soon after, \citet{PaulinICCV15Local} also presented matching results using AlexNet.
Both papers use the Euclidean distance between descriptors and compare matching mean average precisions (mAP) with those for the SIFT descriptor. However, \citet{FischerX14Descriptor} compute descriptors for regions detected with MSER, 
whereas \citet{PaulinICCV15Local} use the Hessian-affine detector.
Noting that AlexNet has five convolutional layers followed by three fully-connected layers, the results of \citet{FischerX14Descriptor} suggest that the performance of AlexNet-based descriptors is nearly independent of the choice of layer if the best input patch sizes are chosen.
Further, the results show that AlexNet-based descriptors clearly outperform SIFT descriptors for a wide range of choice of layer and patch size. In contrast, \citet{PaulinICCV15Local} show a clear preference for using AlexNet's fourth convolutional layer, in line with the results 
of \citet{LongNIPS14Convnets}, arguing that earlier layers of such networks tend to encode more task-independent information.
Moreover, they find higher mAP using SIFT descriptors than AlexNet-based descriptors. \\

Both \citet{FischerX14Descriptor} and \citet{PaulinICCV15Local} propose other descriptors, and demonstrate that they clearly outperform both AlexNet activations and SIFT. 
The descriptors proposed by \citet{FischerX14Descriptor}, which \citet{PaulinICCV15Local} call \emph{PhilippNet}, use the same network architecture as AlexNet, but train it as follows. 
A collection of 16000 random `seed' patches were extracted from Flickr images. Next, 150 random geometric and photometric transformations were applied to each such seed patch. The CNN was trained to associate a single class label to all transformed versions of a given seed patch. 
Meanwhile, \citet{PaulinICCV15Local} proposed a patch-convolutional kernel network (patch-CKN) which exploits a fast and simple stochastic procedure to compute a finite-dimensional feature embedding that approximates a kernel feature map.  
\\

\begin{figure}[ttt]
\begin{center}
\includegraphics[width=\textwidth]{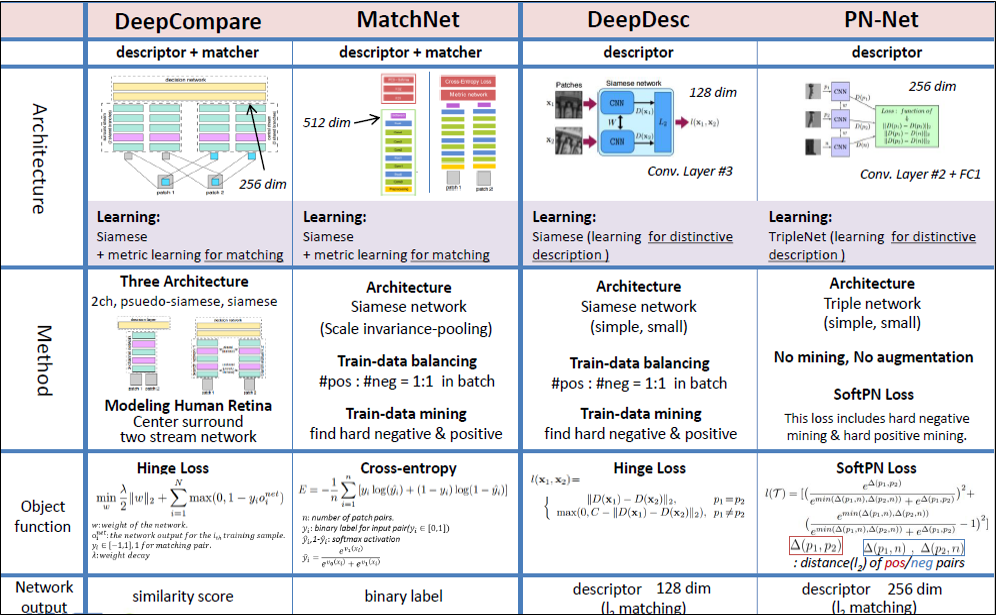}
\caption{An overview of some of the first Siamese CNN-based local features learning methods. From left to right, 
DeepCompare \citep{ZagoruykoCVPR15Learning},  MatchNet \citep{HanCVPR15MatchNet}, DeepDesc \citep{SimoSerraICCV15Discriminative}
and PN-Net \citep{BalntasX16PNNet}. (By courtesy of Yukyung Choi.)} 
\label{fig:deepsummary}
\end{center}
\end{figure}


\textbf{Metric Learning.} 
While nearly all other work discussed in this section learns descriptors and compares them with the Euclidean metric, 
MatchNet\footnote{MatchNet code and pre-trained model available at \url{http://www.cs.unc.edu/~xufeng/matchnet}.}~\citep{HanCVPR15MatchNet} jointly learns a descriptor \emph{and} a metric for comparing descriptors.
It computes descriptors by combining a CNN with multiple convolutional and spatial pooling layers plus an optional
bottle neck layer.
Meanwhile, the metric network passes such descriptors from two patches through three fully-connected layers (see Figure~\ref{fig:deepsummary}).
MatchNet is trained by concatenating two copies of the descriptor network with a metric network and using a cross-entropy loss, thereby transforming the matching problem into a classification problem. 
The authors note that this architecture has similarities with to that in \citet{ZbontarJMLR16Stereo}, which was designed to learn a similarity between patches that is used as a matching cost for a stereo algorithm. Although MatchNet improves matching accuracy, it is not obvious how to combine it with fast approximate nearest neighbour algorithms, like hierarchical navigable small worlds~\citep{MalkovX16Efficient},  which rely on the use of Euclidean distances. \\

\begin{figure}[ttt]
\begin{center}
\includegraphics[width=\textwidth]{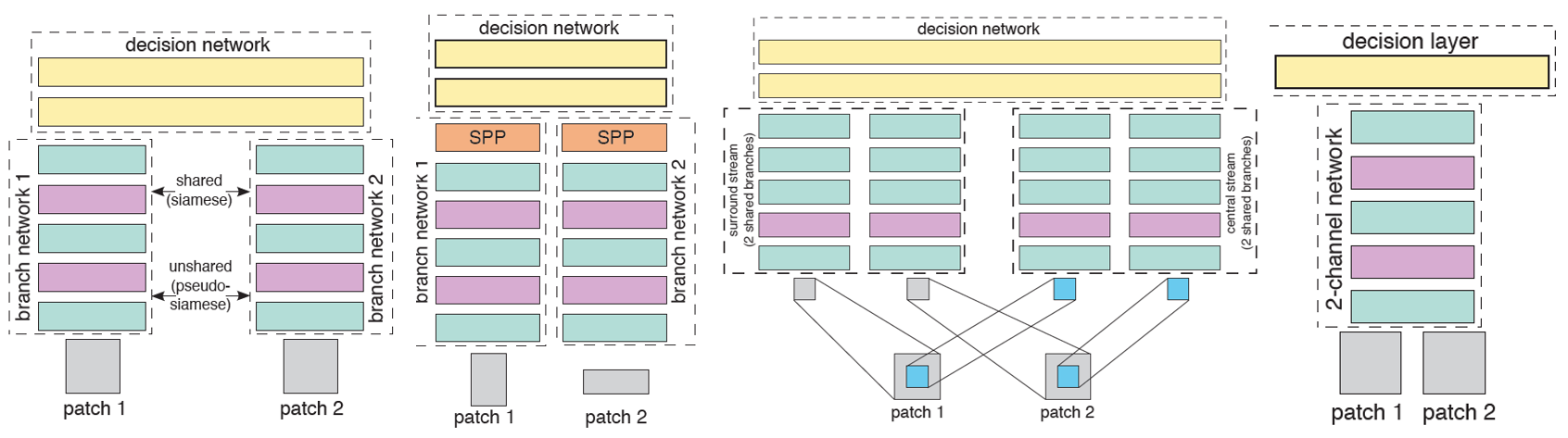}
\caption{Various network architectures explored in \citet{ZagoruykoCVPR15Learning}. From left to right: 
Siamese and pseudo-Siamese networks (the difference between these is that the pseudo-Siamese network does not have shared branches), spatial pyramid pooling (SPP) Siamese network, central-surround two-stream network, and the two-channel network. (By courtesy of Sergey Zagoruyko.)} 
\label{fig:Siamese_archs}
\end{center}
\end{figure}

Similarly, \citet{ZagoruykoCVPR15Learning}\footnote{Source code and trained models are available at \url{http://imagine.enpc.fr/~zagoruys/deepcompare.html}.} also focus on learning metrics for comparing patches.
An interesting aspect of the paper is that it explores a variety of different neural network models for the task, that are collectively known as \emph{DeepCompare}, namely: Siamese networks, pseudo-Siamese networks and two-channel networks, spatial pyramid pooling (SPP) Siamese networks, as well as two-stream multi-resolution models called central-surround two-stream networks (see Figure~\ref{fig:Siamese_archs}).
In contrast to Siamese networks, whose two streams have identical architectures and identical weights, pseudo-Siamese networks have two branches whose architectures are identical but whose weights are \emph{not} shared, allowing for additional flexibility. 
Meanwhile, in a two-channel network, pairs of patches are directly fed to the network as two-channel images and hence they are processed jointly. 
The spatial pyramid pooling (SPP) Siamese network inserts a spatial pyramid pooling layer between the convolutional layers and the fully-connected layers of the network. Such a
layer allows the processing of input patches of different sizes. Finally, the central-surround two-stream network consists of two separate streams: surround, which takes full $64\times 64$ patches as input; and central, which takes only the central $32\times 32$ portions of those patches as input. This enables processing to take place at two different spatial resolutions. 
The paper compares these methods on patch matching and wide-baseline stereo benchmarks, finding that the two-channel network clearly outperforms the others. 
Nevertheless, of the methods evaluated, the two-channel network would be the most computationally expensive in practice if many pairs of patches have to be tested against each other in a brute-force manner. \\


\textbf{Pairwise Losses.}
To compare two patches, most recent methods apply the same CNN to extract a descriptor from each patch and they take the Euclidean distance between these descriptors. All CNNs discussed in the rest of this section operate like this (except~\citet{WeiCVPR18Kernelized} who apply a Gaussian kernel to a ``projection distance''). 
Furthermore, all these CNNs learn from a large set of pairs of patches that are either known to correspond or known to not correspond (see Figure~\ref{fig:deepsummary}). 
However, these CNNs differ in the type of loss used during training. In particular, some use \emph{pairwise losses}, 
where each term involves the distance between two patches, 
some use \emph{triplet losses}, where each term involves the distance between three patches
and others use \emph{global losses}, where each term involves distances between more than three patches.\\

The use of a pairwise loss for learning real descriptors was first proposed by \citet{JahrerCVWW08Learned}, years before the explosive growth in popularity of CNNs in computer vision. In particular, those authors used a square/exponential loss, which is proportional to the squared Euclidean distance $d^2$ for patches corresponding to the same 3D point and proportional to a negative exponential $\exp(-\alpha d)$ for some $\alpha\in\R$, for patches corresponding to different 3D points. \\

DeepDesc\footnote{Torch7 code and pre-trained models for DeepDesc are available at
\url{https://github.com/etrulls/deepdesc-release}.}~\citep{SimoSerraICCV15Discriminative}
learns 128-dimensional descriptors (the same dimensionality as SIFT) whose Euclidean distances reflect patch similarity (see also Figure~\ref{fig:deepsummary}).
Based on the observation that after a certain stage in the learning process, most pairs are correctly classified and no-longer bring an improvement in the performance of the descriptors, they propose a strategy of aggressive mining of ``hard'' positives and negatives. This is done by selecting, after each forward-propagation, the non-corresponding pairs that are hardest to discriminate
and the corresponding pairs that match most poorly.
Only such pairs are then backpropagated through the network. \\

\begin{figure}[ttt]
\begin{center}
\includegraphics[width=\textwidth]{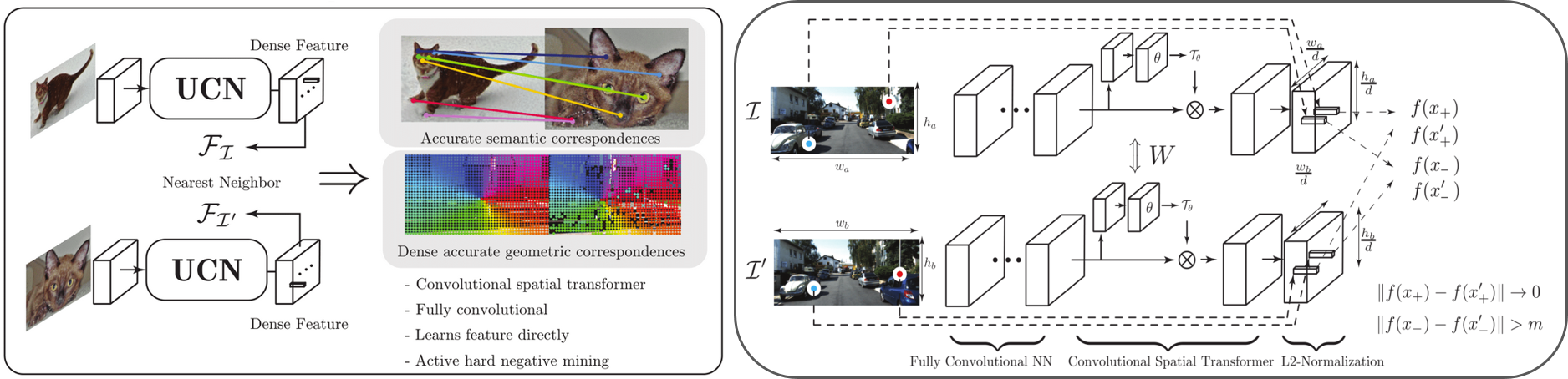}
\caption{The UCN architecture \citep{ChoyNIPS16Universal} which can 
accurately and efficiently learn a metric space for geometric correspondences, dense trajectories or semantic correspondences.
It consists of a fully convolutional neural network combined with a  convolutional spatial transformer and followed by channel-wise $L_2$ normalisation. Features that correspond to the positive points (from both images) are trained to be closer to each other, while features that correspond to negative points are trained to be a certain margin apart. 
(By courtesy of Christopher B. Choy.)} 
\label{fig:UCN}
\end{center}
\end{figure}

\citet{ChoyNIPS16Universal} propose the UCN\footnote{The UCN code, license agreement, and pre-trained models are available at \url{http://cvgl.stanford.edu/projects/ucn/}.}
(Universal Correspondence Network) which also works with Euclidean distance between pairs of $L_2$-normalised descriptors.
Its particularity compared to previous approaches is that it provides a single framework to efficiently handle geometric correspondences, dense trajectories and semantic correspondences. Furthermore, network's input is a whole image, not only an extracted patch, making the method suitable for dense correspondence (see Figure~\ref{fig:UCN}). 
The network extracts descriptors from each image using the initial layers of GoogLeNet~\citep{SzegedyCVPR15Going}. 
It is trained with a \emph{correspondence contrastive loss}, whose value for a given pair of descriptors $f,f'$ for points $x,x'$ from two images 
is proportional to 
\begin{align*}
\begin{cases} \norm{f-f'}_2^2 & \text{if points $x,x'$ correspond} \\
\left( \max\{0, m- \norm{f-f'}_2 \} \right)^2 & \text{if points $x,x'$ do not correspond} \end{cases}
\end{align*}
where $m$ is a hyperparameter.
Inspired by the GPU implementation in \citet{GarciaICIP10Knearest}, the authors implement nearest neighbour search as a Caffe~\citep{JiaX14Caffe} layer to mine hard negatives on-the-fly. 
The authors optionally include a \emph{convolutional spatial transformer} in their network, as proposed by  \citet{JaderbergNIPS15Spatial}. This transformer attempts to make the extracted features invariant to particular families of transformations. The results of the paper suggest that the transformer is beneficial for semantic correspondence tasks or when there are large geometric transformations between images, but otherwise this component can reduce the quality of matching.
\\


\textbf{Triplet Losses.}
\citet{BalntasX16PNNet} proposes the use of a softened version of the triplet loss for training descriptors to be matched with the Euclidean distance\footnote{Code for PN-Net is available at \url{https://github.com/vbalnt/pnnet}.},
calling the resulting network PN-Net, where P and N stand for positive and negative (see Figure~\ref{fig:deepsummary}). 
This work was continued in~\citet{BalntasBMVC16Learning},  
where the authors compare a variety of alternative triplet losses and binary-classification-based losses for descriptor training.
Also, \citet{YangICCV17DeepCD} propose a set of methods called DeepCD\footnote{Code for DeepCD is available at \url{https://github.com/shamangary/DeepCD}.}
which involve augmenting the PN-Net architecture with an extra stream. This extra stream learns a \emph{complementary descriptor}: that is, a descriptor which is intended to help the descriptor output by the PN-Net stream. The authors experiment with complementary descriptors that are either
real vectors compared with Euclidean distance, or binary vectors compared with Hamming distance. The final dissimilarity between two patches is taken as the product of the distance between descriptors from the PN-Net stream and the distance between descriptors from the complementary stream.\\

\citet{MishchukNIPS17Working} propose HardNet\footnote{HardNet code and pre-trained model are available at \url{https://github.com/DagnyT/hardnet}.}, which has the same network architecture as L2-Net, but uses a much simpler loss. This loss maximises the difference in Euclidean distance between the closest positive and closest negative example in each batch and is formulated as follows. Each batch $i = 1, 2, \dots, n$ has exactly one pair of patches corresponding to the same 3D point, whose descriptors are $a_i$ and $p_i$.
Also, batch $i$ has two sets of patches that do not correspond, whose descriptors form the sets $\mathcal{A}_i$ and $\mathcal{P}_i$. 
The loss is then
\begin{align*}
L = \frac{1}{n} \sum_{i=1}^n \max\left\{ 0, 1 + \norm{a_i-p_i}_2 - \min\{ \min_{p\in\mathcal{P}_i} \norm{a_i-p}_2,  \min_{a\in\mathcal{A}_i} \norm{a-p_i}_2 \} \right\}.
\end{align*}
The authors explore the effect of batch size on performance, finding that false positive rates decline with increasing batch size until the batch size reaches 512 after which performance saturates.
The use of this loss can be seen as a local hard negative mining strategy. Note also that the way the triplets are formed does not require the network to be three-streamed, in contrast to other triplet-based approaches such as TFeat~\citep{BalntasBMVC16Learning}. \\

\citet{WeiCVPR18Kernelized} propose a new pooling method, called subspace pooling (SP), which enables invariance to a range of geometric deformations such as circular shifts, flipping and in-plane rotation.
The basic idea of SP is to model the convolutional feature maps as the linear subspace spanned by their principal components. The authors show that the proposed SP is invariant to all the geometric changes that can be expressed as column permutations of the matrix formed by the feature maps from the last convolution layer stacked as one dimensional features. SP is therefore similar to the bilinear pooling function in \citet{LinICCV15Bilinear}, which is equivalent to spatial reordering.
Integrating the proposed pooling method with models such as HardNet or TFeat, substantially improves matching results for those models. \\


\textbf{Global Losses.}
TGLoss\footnote{Matlab code and  pre-trained models are available at \url{https://github.com/vijaykbg/deep-patchmatch}.}~\citep{KumarCVPR16Learning} is another triplet network, in the sense that the network is applied to three patches at a time during training. However, it replaces the triplet loss with a \emph{global} loss. This global loss involves the mean $\mu_+$ and variance $\sigma_+^2$ of the distribution of Euclidean distances between descriptors of matching patches, as well the equivalent mean $\mu_-$ and variance $\sigma_-^2$ for non-matching patches.
The intent is to keep the sum of variances $\sigma_+^2+\sigma_-^2$ small while ensuring that the different of the means $\mu_- - \mu_+$ is large. 
\\

Similarly to DeepDesc, L2-Net\footnote{A Matlab implementation of L2-Net with pre-trained models is available at \url{https://github.com/yuruntian/L2-Net}.}~\citep{TianCVPR17L2Net} also uses CNNs to learn high-performance
128-dimensional descriptors whose Euclidean distances reflect patch similarity. 
However, L2-Net has a CNN with 7 convolutional layers, a final layer that normalises the outputs to unit Euclidean length, and batch normalisation is employed during training, whereas DeepDesc's network has only 3 convolutional layers and no batch normalisation is used.
Also, L2-Net abandons DeepDesc's idea of mining hard samples and instead uses batches consists of $p:=128$ correctly-matching pairs of patches plus $p^2-p$ non-matching pairs from the same patches. This better models the rate of correct and incorrect matches to be expected in real data than the triplet loss usually applied to a Siamese network. 
Also, L2-Net uses a loss function consisting of three terms: one accounts for the relative distance between descriptors, one controls descriptor compactness, and the other is an extra supervision imposed on the intermediate feature maps. 
Finally, as the output of L2-Net approximates a zero-mean Gaussian distribution, the authors find that by applying the sign function to this output they obtain a highly-effective set of binary descriptors. \\

The main idea of global orthogonal regularisation (GOR)\footnote{A TensorFlow implementation is available at \url{https://github.com/ColumbiaDVMM/Spread-out_Local_Feature_Descriptor}.}~\citep{ZhangICCV17Learning} is to force features to be ``spread-out'' in the descriptor space in order to fully utilize the expressive power of that space. This regulariser encourages randomly-sampled non-matching descriptors to resemble uniformly-distributed points on the unit sphere embedded in $\R^d$, where $d$ is the dimension of the descriptor space. It does so by forcing the sample mean and second moment of the inner product of the descriptors of non-matching pairs to be close to zero and $1/d$ respectively. The authors show that adding this loss to models such as DeepDesc or TFeat (discussed above) results in better patch matching. \\


\textbf{Histogram Losses.}
Most of the systems discussed above, including PN-Net, L2-Net and HardNet, are \emph{trained} to minimise a pairwise or triplet loss, with or without some hard positive or negative mining, but they are \emph{evaluated} in terms of mean average precision (mAP), which is the area under the precision-recall curve. Would one not expect higher mAP if one were to train such systems to directly maximise mAP?
\citet{HeCVPR18Local} propose DOAP (descriptors optimised for average precision) which does exactly that, 
using the same network architecture as L2-Net and HardNet.
The authors propose methods for generating both real-valued and binary descriptors. 
It is not immediately obvious how to effectively approximate the mAP with a differentiable objective function.
The authors do so based on results from their own paper~\citep{HeCVPR18Hashing} on hashing and learning binary embeddings,
which are based on the idea of building losses using histograms~\citep{UstinovaNIPS16Learning}.
In particular, they compute a histogram of the Euclidean distances between a query descriptor and all descriptors in a given minibatch, treating all descriptors in a given bin as ``ties'' which have an equal distance from the query descriptor.
They then plug this histogram into an efficient tie-aware formulation of average precision from~\citet{McSherryECIR08Computing}. 
To make this formulation of average precision differentiable, they replace the discrete histogram counting operation by partially assigning each inter-descriptor distance to its two closest histogram bins with linear interpolation. \\

While the basic DOAP descriptors already demonstrate state-of-the-art performance on standard benchmarks as we discuss in Section~\ref{sec:discuss}, the authors also experiment with two improvements.
The first improvement (DOAP-ST) is to add a spatial transformer module \citep{JaderbergNIPS15Spatial} to make matching more robust to challenging levels of geometric noise and illumination change. 
The second improvement (DOAP-LM) is to use \emph{label mining}, which is a clustering method that avoids forcing the system to discriminate between visually similar patches that correspond to distinct 3D points, such as patches from different windows of the same style on the same building.\\

\subsection{Learning binary descriptors with CNNs}
\label{sec:deepbin}

Having seen some methods for learning binary descriptors
in Section~\ref{sec:learn_binary}, it is not surprising that deep-learning methods have also been designed to tackle this problem. 
Indeed, some of the methods discussed in Section~\ref{sec:deepdesc}, such as L2-Net~\citep{TianCVPR17L2Net}, also have binary variants. 
However, such methods essentially apply the sign function to each component of a real descriptor vector, and better-performing alternative methods have been proposed that are specifically designed to learn binary features.
\\

DeepBit\footnote{Code available at \url{https://github.com/kevinlin311tw/cvpr16-deepbit}.}~\citep{LinCVPR16Learning} is an unsupervised deep-learning approach to learning compact binary descriptors for efficient visual object matching. The main idea is to optimise the parameters of a network using a combination of three losses.
The first loss forces the binary descriptors to preserve the local data structure
by minimising the quantisation loss when the activations of the last layer are projected into binary descriptors.
The second loss encourages each bit of the binary descriptor to be evenly distributed, with the intention of making the descriptor more discriminative.
Finally, the third loss encourages the descriptor to be invariant to rotations, simply by penalising changes in the descriptor if the input patch is rotated.  \\

DBD-MQ (deep binary descriptor with multi-quantisation)~\citep{DuanCVPR17Learning} is another unsupervised descriptor learning method. The authors propose a novel and effective approach to quantising real descriptors that they call a $K$-autoencoder network.
This consists of $K$ autoencoders, each of which uses a $c$-bit binary representation of a given real descriptor. They experiment with $c = 16, 32$ and $64$ and find that $K=4$ gives the best mAP on a matching task. 
The binary descriptor is the result of concatenating the $K$ binary representations.
The autoencoders are trained by analogy with the $k$-means clustering algorithm:
given a real descriptor, they find the autoencoder giving the least reconstruction error; then they backpropagate the error associated with that descriptor through that autoencoder alone. \\

\subsection{End-to-end detection and description of local features}
\label{sec:deep}

Relative to the number of deep descriptors, only a few end-to-end deep models have been proposed that aim to learn the entire detection and description pipeline (Figure~\ref{fig:locfeat}). This is in spite of the fact that most handcrafted detectors rely on convolutional filters just like CNNs, as well as the fact that one motivation for deep learning is to work directly with raw input rather than relying on the output of handcrafted (and therefore suboptimal) preprocessing.
\\

\begin{figure}[ttt]
\begin{center}
\includegraphics[width=0.95\textwidth]{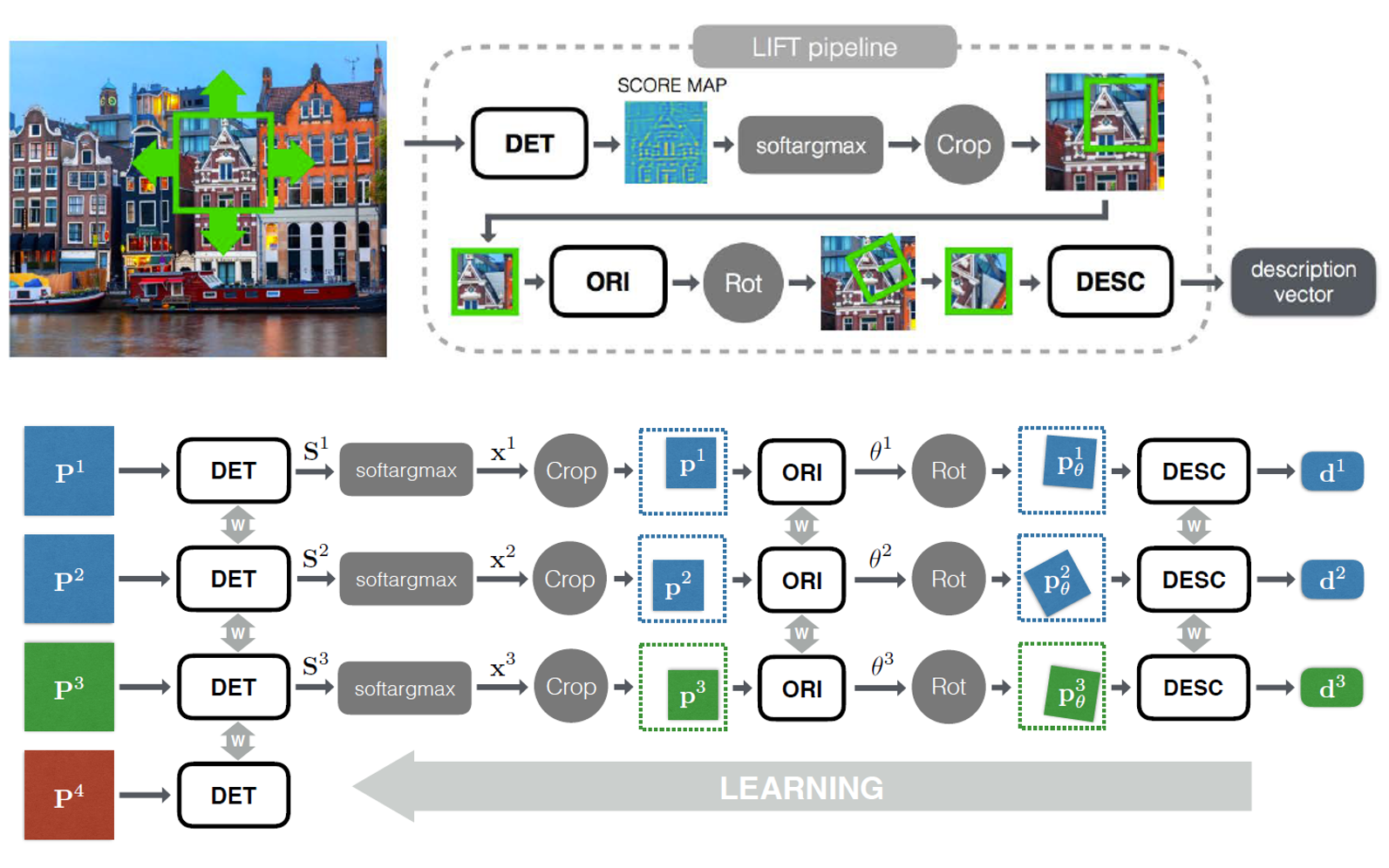}
\caption{Top: The LIFT pipeline from~\citet{YiECCV16LIFT}. Given an input patch $\bfP$, LIFT applies a detector (DET) 
followed by a softargmax and a crop operation to provide a smaller patch $\bfp$ that lies within $\bfP$.
The smaller patch is fed to an orientation estimator (ORI) and rotation operation which provides rotated patches as input for 
a network that computes descriptors (DESC). 
Bottom: LIFT's Siamese training architecture that
works with quadruplets of patches. Patches $\bfP^1$ and $\bfP^2$ (blue) correspond to different views of
the same physical point and are used as positive examples to train the descriptor (DESC).
Patch $\bfP^3$ (green) shows a different 3D point which serves as a negative example for DESC.
Finally patch $\bfP^4$ (red) contains no distinctive feature points and is only used as a
negative example to train the detector (DET). (By courtesy of Kwang Moo Yi.)}
\label{fig:LIFT}
\end{center}
\end{figure}

\citet{YiECCV16LIFT} appear to have been the first authors to propose a fully end-to-end approach to feature detection and description.
Their LIFT\footnote{Code and pre-trained models are available at \url{https://github.com/cvlab-epfl/LIFT}.} (learned invariant feature transform) network is composed of three CNN components that feed into each other (see Figure~\ref{fig:LIFT}): a detector, an orientation estimator and a descriptor. These components are linked with two spatial transformers~\citep{JaderbergNIPS15Spatial} that geometrically manipulate the image patches while preserving differentiability. 
The first spatial transformer crops the input patch to give a smaller patch centred at the point output by the detector, while the second rotates this smaller patch to the angle determined by the orientation estimator. 
The authors use the TILDE detector~\citep{VerdieCVPR15TILDE}. 
At training time, they modify TILDE to ensure differentiability by replacing non-maximum suppression with a softargmax function, which is defined as follows. If $S(y)\in\R$ notes the score for image coordinate $y$ output by TILDE for a patch $P$ with domain $\text{dom}(P)$, then the estimated location of the feature point is 
\begin{align*}
\text{softargmax}(S) := \frac{\sum_{y\in\text{dom}(P)} y e^{\beta S(y)}}{\sum_{y\in\text{dom}(P)} e^{\beta S(y)}}
\end{align*}
where $\beta\in\R$ is a hyperparameter.
The orientation estimator relies on the CNN proposed by \citet{YiCVPR16Learning} which was designed to estimate orientations for matching purposes\footnote{Code is available at \url{https://github.com/cvlab-epfl/learn-orientation}.} and which demonstrates significant improvements over SIFT's orientation estimator.
Finally, DeepDesc~\citep{SimoSerraICCV15Discriminative} is used for the descriptor component, since it is a simple network, which 
does not require metric learning.
The authors found it impossible to learn the entire network from scratch. Instead, they initialise the weights by first learning the descriptor, then using that descriptor to learn the orientation estimator, and then using the orientation estimator and descriptor to learn the detector. 
Finally, the whole network is trained end-to-end with quadruplets of patches, as shown in Figure~\ref{fig:LIFT}. 
At test time, scaled versions of the image are fed to the network, and the detector generates score maps at each scale, which are processed by non-maximum suppression to give keypoint locations. The orientation detector and descriptor are then applied only to patches at the detected keypoint locations. \\
 
In contrast to the other end-to-end models discussed here, DELF\footnote{DELF code available at 
 \url{https://github.com/tensorflow/models/tree/master/research/delf }.} (deep local features) \citep{NohICCV17LargeScale} 
was designed to perform more accurate matching and geometric verification for large-scale image retrieval.
The authors use visual attention for keypoint selection, arguing that keypoint selection is important for both accuracy and computational efficiency of retrieval systems, since a substantial fraction of local features are irrelevant and may distract such systems.
While visual attention based on deep neural networks had previously been employed for many other computer vision tasks~\citep{HongNIPS15Decoupled,XuICML15Show}, it had not previously been used for image retrieval.
The DELF pipeline has four main blocks: (i) dense feature extraction, using an intermediate output of a ResNet50~\citep{HeCVPR16Deep}; (ii)
attention-based keypoint selection, using a two-layer CNN with a softplus activation at the top (softplus is the function $x\mapsto \log(1+\exp(x))$);
(iii) dimensionality reduction, using PCA; and (iv) indexing and retrieval, in which 40 local features from each query image are matched with a database by nearest-neighbour search and the matches are geometrically verified with RANSAC.
The results show that DELF significantly outperforms pre-existing methods for image retrieval in terms of its precision-recall curve.
In particular, unlike other methods evaluated, it is robust to queries that have no match in the database.\\

\begin{figure}[ttt]
\begin{center}
\includegraphics[width=\textwidth]{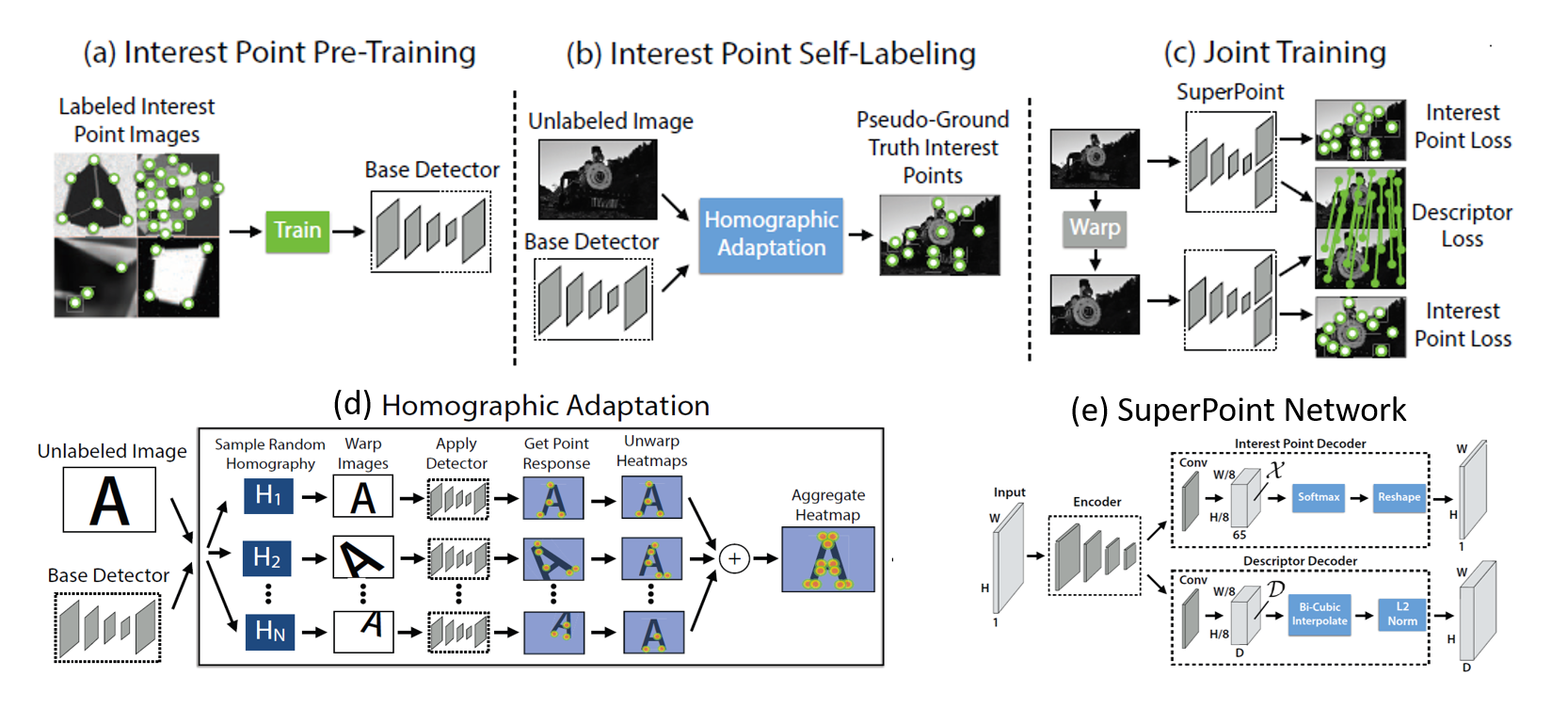}
\caption{Overview of the SuperPoint framework for self-supervised training of keypoint detectors from \citet{DeToneDLVS18SuperPoint}. 
\emph{Keypoints} are here referred to as \emph{interest points}, consistent with the authors of that paper.
(a) An initial interest point detector is pre-trained using synthetic data. 
(b) Then the homographic adaptation procedure automatically generates pseudo-ground truth interest points. 
(c) The pseudo-ground truth interest points are used to train the SuperPoint network
that jointly extracts interest points and descriptors from an image.
(d) Homographic adaptation consists in averaging the response of a detector over a set of random homographies. 
(e) The SuperPoint network consists of a single encoder whose output is fed to two decoders. 
(By courtesy of Daniel DeTone.)} 
\label{fig:SuperPoint}
\end{center}
\end{figure}

SuperPoint~\citep{DeToneDLVS18SuperPoint} is a framework for
self-supervised training of keypoint detectors and descriptors for multiple-view geometry problems. 
Given an input image, it jointly computes keypoint locations and descriptors in a single forward pass. 
As shown in Figure \ref{fig:SuperPoint}, first a base keypoint detector is trained on examples from a synthetic dataset consisting of 
simple geometric shapes for which the keypoint locations are well defined. Then, to improve the repeatability of the detector and enlarge the set of stimuli to which it responds, a process called \emph{homographic adaptation} is applied, which essentially averages the output of a detector over a suitably-chosen distribution of random homographies. 
Specifically, the authors define the homographic adaptation $f^\text{HA}$ of detector $f$ applied to image $I$ as the average
\begin{align*}
f^\text{HA}(I) := \frac{1}{n} \sum_{i=1}^n H_i^{-1}(   f( H_i(I) )   )
\end{align*}
where $(H_i)_{i=1}^n$ is a sequence of homographies.
The keypoints detected by the homographic adaption of the base keypoint detector, which the authors call \emph{pseudo-ground truth interest point locations}, are collected and used as training data for the SuperPoint network.
This network has a single encoder whose output is fed into a pair of decoders.
One decoder performs interest point detection and the other computes descriptors.
The detector is trained with a cross-entropy loss, where labels are derived from the pseudo-ground truth interest point locations.
Meanwhile, the descriptor is trained using the sum of a hinge loss for pairs of descriptors that are known to correspondent
and another hinge loss for pairs of descriptors that are known to not correspond.\\

\section{Performance comparisons}
\label{sec:discuss}

In this section, we begin by discussing performance comparisons of handcrafted local features on matching and patch retrieval tasks (Section~\ref{sec:handcrafted-pc})
and move on to comparisons that also involve deep-learning-based features (Section~\ref{sec:deep-pc}). Finally, we discuss an evaluation of the impact of the choice local features on the complex task of image-based reconstruction (Section~\ref{sec:recon-pc}).

\subsection{Comparisons of handcrafted local features}
\label{sec:handcrafted-pc}
\citet{HeinlyECCV12Comparative} analysed the performance of different handcrafted detector and descriptor pairings on several datasets, namely the Harris, MSER, FAST, BRIEF, ORB, BRISK, SURF and SIFT detectors, and the BRIEF, ORB, BRISK, SURF and SIFT descriptors.
The authors examined the impact of both geometric and non-geometric changes (blur, JPEG compression, exposure, day-to-night) on matching. 
They defined the \emph{matching score} as the ratio of the number of correct matches to the number of features detected in the first image to be matched. 
In terms of matching scores and in the presence of non-geometric changes, the results showed that BRIEF descriptors beat ORB, BRISK and even SIFT descriptors, and that they did so whichever detector was used. 
However, as soon as there was any rotation between the images to match, BRIEF descriptors performed awfully compared with ORB and BRISK descriptors.
In general, in the presence of geometric changes, the matching scores for the SIFT detector and descriptor were consistently better than those for the other detector-descriptor pairs. 
\\
 
\begin{figure}[ttt]
\begin{center}
\includegraphics[width=\textwidth]{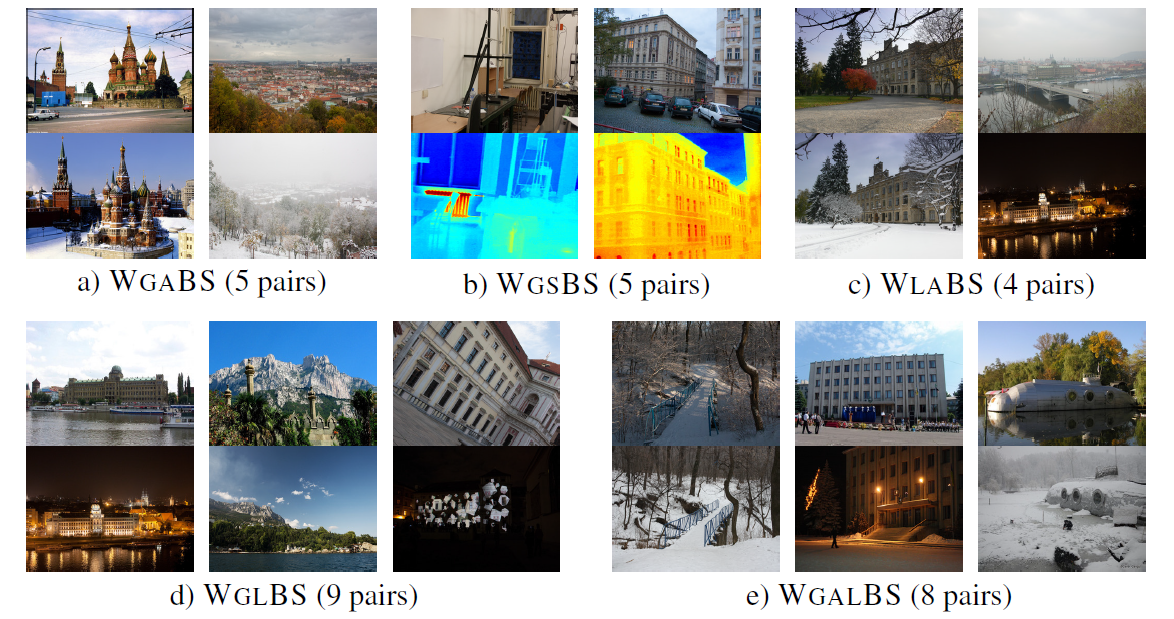}
\caption{Example images from the  \textsc{WxBS} wide-baselines dataset proposed by \citet{MishkinBMVC15WxBS}.
\textsc{WgaBS} stands for viewpoint and appearance changes, 
\textsc{WgsBS} for viewpoint and modality changes,  
\textsc{WlaBS} for lighting and  appearance changes, 
\textsc{WglBS} for viewpoint and  lighting changes, and 
\textsc{WgalBS} for viewpoint, appearance and lighting changes. (By courtesy of Dmytro Mishkin.)} 
\label{fig:WXBS}
\end{center}
\end{figure}

\citet{MishkinBMVC15WxBS} propose the wide multiple baseline stereo (\textsc{WxBS}) dataset\footnote{The WxBS dataset can be found at \url{http://cmp.felk.cvut.cz/wbs/index.html}.}, which consists of image pairs with a variety of combinations of large changes in geometry, illumination, sensor, appearance and modality. Every image pair is manually labelled with approximately 20 correspondences as ground-truth.  
The authors explore matching visible-spectrum images with infrared images, matching images acquired with different modes of magnetic resonance imaging, and even matching maps to satellite views (see Figure \ref{fig:WXBS}). They show that this is an extremely challenging benchmark and that using only one detector-descriptor model at a time does not perform well. 
To quantify matching performance, the authors compare the number of images for which at least 15 correct inliers to a homography are found. 
Of the detector-descriptor combinations that the authors explore\footnote{14 such combinations are explored in the poster associated with the paper, which can be found at~\url{http://cmp.felk.cvut.cz/~mishkdmy/posters/wxbs-2015-poster.pdf}, but only up to 13 combinations were tested, depending on the dataset, in the paper itself.},
the best results are obtained with models that incorporate multiple detectors and descriptors, such as those proposed by \citet{YangPAMI07Registration} and \citet{MishkinX15MODS}.
However, in terms of single-detector/descriptor approaches, an adaptive Hessian-affine detector was the best-performing detector,
while SIFT and its variants including DAISY~\citep{TolaPAMI10Daisy} were the best-performing descriptors.  
Furthermore, it was observed that most of the descriptors gain significantly from photometric normalisation. 
One of the main problems in day-to-night matching and matching infrared images is the low number of detected features. 
A possible approach addressing this problem is iiDoG~\citep{VonikakisMST13Biologically}, where the SIFT detector's difference of Gaussians is normalised by sum of Gaussians, but this approach cannot be easily be applied to other detectors. \\

\citet{MaierCVPR17Ground} presented a method to generate ground-truth matches based on the original ground truth of well-known
datasets, pointing out that previous notions of a ``correct match'' were ambiguous, partly because they relied on arbitrarily-set thresholds.
For instance,~\citet{MikolajczykIJCV05Comparison} used a threshold of $40\%$ on an ``overlap error criterion'',
\citet{MikolajczykPAMI05Performance} used a threshold $50\%$ on an ``overlap error criterion'' whereas \citet{HeinlyECCV12Comparative} used a threshold of 2.5 pixels on the location of keypoints.
The authors also conducted extensive tests on 133 keypoint-descriptor combinations on the HCI Training 1K flow~\citep{KondermannCVPR16The}, KITTI~\citep{MenzeCVPR15Object} and Oxford~\citep{MikolajczykPAMI05Performance} datasets.
The best average accuracy was attained by the combination of the BRISK~\citep{LeuteneggerICCV11Brisk} detector and the FREAK~\citep{AlahiCVPR12Freak} binary descriptor, with an average accuracy of $81.5\%$, closely followed by the combination of the MSD~\citep{TombariACCV14Interest} detector and the optimised ConvOpt~\citep{SimonyanPAMI14Learning} descriptor.
Meanwhile, the LATCH~\citep{LeviX15LATCH} and BOLD~\citep{BalntasCVPR15Bold} descriptors performed poorly no matter which detector they were paired with, having an average accuracy consistently under 60\%.\\

\begin{figure}[ttt]
\begin{center}
\includegraphics[width=\textwidth]{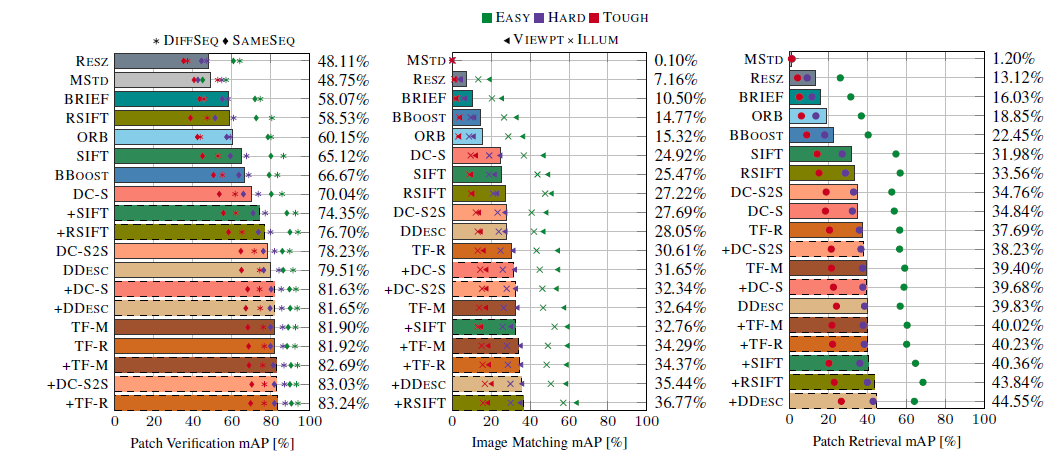}
\caption{Mean average precisions (area under the curve) for patch verification, image matching and patch retrieval tasks in the HPatches benchmark~\citep{BalntasCVPR17HPatches}. Before extracting descriptors, each patch is randomly perturbed by applying a rotation, translation and anisotropic scaling, each of which is drawn from a uniform distribution. The colours of the markers indicate the ``difficulty'' of this geometric-transformation noise, which corresponds to the ranges of the uniform distributions. Triangular markers correspond to results from image sequences with viewpoint changes, while crosses correspond to image sequences with 
photometric changes. Bars show the average of the six variants of each task. Bars with dashed borders and with a $+$ before the descriptor name indicate ZCA projected and normalised features.
(By courtesy of Vassileios Balntas.)}
\label{fig:HPATCH_results}
\end{center}
\end{figure}

\citet{SunICCV17Dataset} propose a dataset that was acquired in a shopping mall for benchmarking image-based localisation algorithms.
The authors compare the localisation performance of BRIEF, SURF, SIFT, COV and RSIFT local features.
COV is the authors' name for the detector proposed by~\citet{PerdochCVPR09Efficient}, which is like the Hessian-affine detector~\citep{MikolajczykIJCV04Scale} but for two modifications. The first modification is to use the scale-space maxima of the Hessian operator for the initial scale selection. The second modification is that in place of computing a rotation for each patch, they use a \textit{gravity vector}~\citep{PhilbinCVPR07Object}, which corresponds to finding the vanishing point of vertical lines in an image. 
Meanwhile RSIFT~\citep{ArandjelovicCVPR12Three}, also known as RootSIFT, is a simple variant of SIFT in which each component $x_i$ of a 
SIFT descriptor $x\in\R^{128}$ is replaced by its square root $\sqrt{x_i}$, noting that these components are non-negative by definition.
\citet{SunICCV17Dataset} measure performance in terms of the fraction of all query images whose camera position is estimated within a given distance from the ground truth \emph{and} whose angle is estimated to within $5^\circ$.
The best performance was obtained with the affine-covariant COV detector coupled with RSIFT descriptors. 
Of course, these results are for a shopping mall environment where the gravity vector is usually well defined and one would not expect the COV detector to be the best choice for environments with less man-made constructions.
\\

\subsection{Comparisons involving deep-learning-based local features}
\label{sec:deep-pc}
In contrast to the papers just discussed, recent performance comparisons have tended to consider both handcrafted \emph{and} deep-learning-based local features.
Such comparisons are the focus of this section. 
In particular, we discuss the work of~\citet{ZhangCVPR17Learning}, who compared handcrafted and deep-learning-based detectors, of~\citet{BalntasCVPR17HPatches}, who introduced a new large new dataset called HPatches for evaluating descriptors from the perspective of patch verification, matching and retrieval tasks, as well as discussing three papers \citep{WeiCVPR18Kernelized,HeCVPR18Local,LencECCV18Large} that made extensive use of the HPatches data. \\

\citet{ZhangCVPR17Learning} compare several handcrafted feature detectors with FAST~\citep{RostenECCV06Machine}, TILDE~\citep{VerdieCVPR15TILDE}, CovDet~\citep{LencX16Learning}
and TCovDet~\citep{ZhangCVPR17Learning} on multiple datasets\footnote{The datasets can be downloaded from \url{https://www.dropbox.com/s/l7a8zvni6ia5f9g/datasets.tar.gz}.}.
The authors show that the repeatability TCovDet keypoints 
is significantly higher than the repeatability of CovDet and TILDE keypoints.
CovDet and TILDE keypoints in turn have higher repeatability than handcrafted methods (namely SIFT, SURF, MSER, Harris Laplace, Hessian Laplace, Harris affine and Hessian affine detector) and FAST. 
In terms of matching scores, TCovDet performs best on two out of three datasets and SIFT on the third dataset, which contains drastic background clutter changes that seem to be handled less-well by TCovDet.  \\

\begin{figure}[ttt]
\begin{center}
\includegraphics[width=\textwidth]{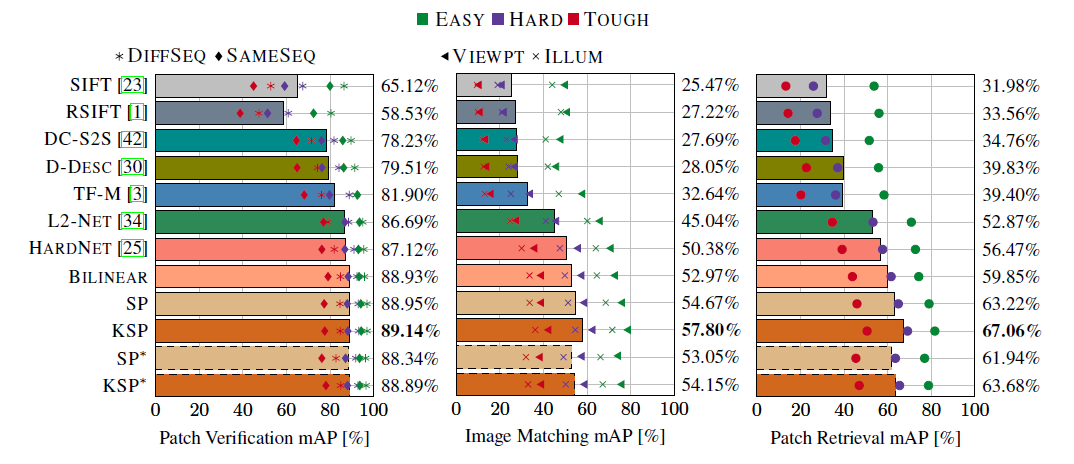}
\caption{Patch verification, matching, and retrieval results from on the HPatches dataset from \citet{WeiCVPR18Kernelized}. The same notation is used as in Figure~\ref{fig:HPATCH_results}. (By courtesy of Xing Wei.)}
\label{fig:KSP}
\end{center}
\end{figure}
 
\citet{BalntasCVPR17HPatches} show that previous datasets (see Table~\ref{tab:patchDataset}) and protocols for evaluating local features do not unambiguously specify all aspects of evaluation, leading to inconsistencies in results reported in the literature. To overcome these weaknesses, the authors propose a new benchmark called HPatches.
This benchmark includes a large new dataset suitable for training and testing modern descriptors.
It also unambiguously defines evaluation protocols for several tasks such as matching, retrieval and classification. The authors conducted an exhaustive evaluation comparing the handcrafted features SIFT, RSIFT\footnote{Using a square root (Hellinger) kernel instead of the standard Euclidean distance to measure the similarity between SIFT descriptors.}~\citep{ArandjelovicCVPR12Three}, BRIEF, ORB, BBoost~\citep{TrzcinskiPAMI15Learning}, as well as the
recent deep descriptors DeepCompare (DC)~\citep{ZagoruykoCVPR15Learning}, DeepDesc~\citep{SimoSerraICCV15Discriminative} and TFeat (TF)~\citep{BalntasBMVC16Learning}.
They also proposed two baseline descriptors, MSTD which is the mean and the standard deviation of the patch, and RESZ which is a vector obtained by resizing the patch to $6\times 6$ and normalising it to have zero mean and unit variance. The results from \citet{BalntasCVPR17HPatches} are shown in Figure~\ref{fig:HPATCH_results}.
The learning-based descriptors were trained on the PhotoTourism~\citep{WindercCVPR07Learning} dataset, which is different from HPatches. Evaluation was done on three benchmark tasks: patch verification, image matching and patch retrieval. The authors also consider applying zero-phase component analysis (ZCA)~\citep{BellNIPS97Edges}, which corresponds to multiplying a descriptor vector by $C^{-1/2}$ where $C$ is the covariance matrix of all descriptor vectors. In most cases, post-processing the descriptors by applying ZCA, followed by power law normalisation~\citep{ArandjelovicCVPR12Three} and $L_2$-normalisation significantly improved the results, as is apparent in Figure~\ref{fig:HPATCH_results} where $+$ indicates use of ZCA. This improvement had already been observed in previous papers \citep{KeCVPR04More,ArandjelovicCVPR12Three}. \\

On the patch verification task in HPatches, deep-learning-based descriptors gave the best mean average precision (mAP). The authors argue that this is because they were jointly optimised with their distance metric to perform well in the verification task. On the image matching task, where descriptors are used to match patches from a reference image to a target image, ZCA whitened and normalised RSIFT surprisingly outperformed the deep-learning-based descriptors. ZCA whitened and normalised RSIFT had almost the highest mAP on the patch retrieval task.
The authors observed that binary descriptors have a competitive mAP only for patch verification. In particular, the learning-based binary feature BBoost in general outperformed the handcrafted binary features especially on the patch verification task.
Among the deep features, the best matching and retrieval performance was obtained with DeepDesc, followed by TF. However, DeepDesc performed worse on patch verification and of the deep-learning-based methods it has the highest computational cost.\\

\begin{figure}[ttt]
\begin{center}
\includegraphics[width=\textwidth]{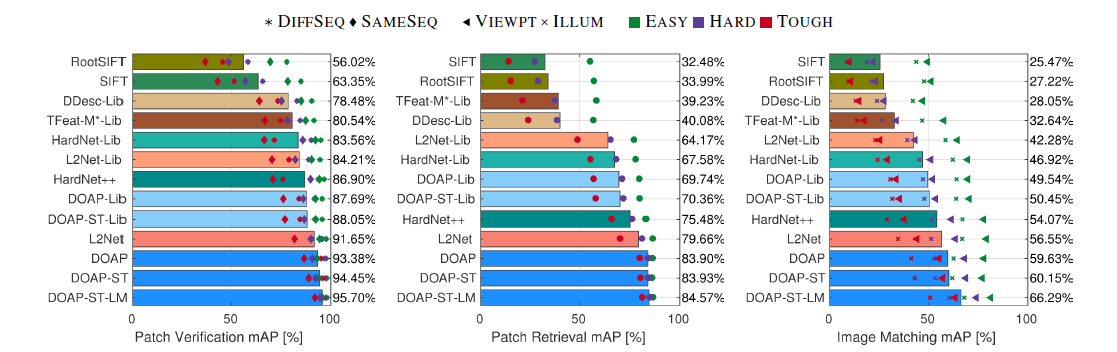}
\caption{Patch verification, matching, and retrieval results on the HPatches dataset from \citet{HeCVPR18Local}. The suffix -Lib means that
the model was trained on the Liberty set from the Photo-Tourism dataset \citep{WindercCVPR07Learning} instead of HPatches, -ST means that the model includes a spatial transformer network to estimate the geometric transformation, and -LM means that clustering-based label mining was used during training. 
The same notation is used as in Figure~\ref{fig:HPATCH_results}. (By courtesy of Kun He.)}
\label{fig:DOAP}
\end{center}
\end{figure}

Several recent papers have used the HPatches benchmark dataset and protocol to evaluate other descriptors. These include \citet{WeiCVPR18Kernelized}, whose results are shown in Figure~\ref{fig:KSP}, as well as \citet{HeCVPR18Local} whose results are shown in Figure~\ref{fig:DOAP}. 
In Figure~\ref{fig:KSP}, the highest mAP is attained by the subspace pooling (SP) descriptors (as described in Section~\ref{sec:deepdesc}) and kernelised subspace pooling (KSP) descriptors (in which the marginal triplet loss was combined with a Gaussian kernel).
However, these are closely followed by \textsc{Bilinear}, which is the model of \citet{LinICCV15Bilinear} applied for learning patch matching. 
Clearly, more recent deep-learning-based approaches such as L2-Net~\citep{TianCVPR17L2Net} and  HardNet~\citep{MishchukNIPS17Working} significantly outperform the previous results on HPatches as shown in Figure~\ref{fig:HPATCH_results}. Furthermore, adding bilinear or subspace pooling to L2-Net further improves performance on all tasks.  \\

Figure~\ref{fig:DOAP} compares variants of DOAP \citep{HeCVPR18Local} with SIFT and other deep models. 
These results are somewhat puzzling in the light of the Figure~\ref{fig:KSP}, since the mAPs for L2-Net differ by large amounts between the two figures.
Assuming that both results are correct, it appears that the SP and KSP descriptors of~\citet{WeiCVPR18Kernelized} are outperformed on every task by DOAP:
using the average precision as loss indeed appears to improve the mAP on patch retrieval, image matching and patch verification tasks.
Furthermore, the two improvements to DOAP (spatial transformer and label mining) consistently
improve the mAP. Notably, label mining increases mAP for image matching by over $6\%$.
These results make DOAP-ST-LM the best-performing descriptor on the HPatches benchmark at the 
time of writing.
\\

Unlike the above performance comparisons that focus on descriptors or detector-descriptor combinations, \citet{LencECCV18Large} focus only on detectors.
In particular, the authors propose an improved repeatability measure for detector evaluation. The first improvement is to correct an error in the publicly-available implementation of the repeatability score of~\citet{MikolajczykIJCV05Comparison}. This error causes the repeatability score to vary strongly if the detected regions are simply scaled (magnified)
by a common factor, even though the score had been explicitly designed to minimise such variation. 
This error concerned a heuristic used for accelerating ellipse overlap computation, which had been applied at the wrong stage of the processing pipeline. 
The second improvement reduces the dependency of the repeatability score on the number of regions detected, which is a tuneable parameter of most detectors. This improvement consists in reporting the average of the repeatability scores attained when 100, 200, 500 and 1000 keypoints are detected per image.\\

\citet{LencECCV18Large} also evaluate 11 existing detectors on 5 datasets using the improved repeatability measure.
One of these datasets, which the authors call \emph{HSequences}, consists of 580 image pairs drawn from the same set of 696 images as the HPatches dataset, making it the largest detector-evaluation dataset at the time of writing. 
In the presence of large illumination changes, they find that the learning-based detector TILDE~\citep{VerdieCVPR15TILDE} has a consistently higher repeatability than the other detectors considered.
However, in the presence of viewpoint changes, TILDE performs poorly since it aims only for translation invariance and not for scale or affine invariance. 
Rather, given viewpoint changes, the Hessian affine detector tends to have the highest repeatability although it is often outperformed by 
CovDet~\citep{LencX16Learning} and TCovDet~\citep{ZhangCVPR17Learning}.\\

\subsection{Comparison on image-based reconstruction task}
\label{sec:recon-pc}

The comparative evaluation proposed in \citet{SchonbergerCVPR17Comparative} goes beyond descriptor
matching, to also evaluate the performance of various descriptors on image-based reconstruction
tasks using challenging small- and large-scale datasets. 
Such image-based reconstruction pipelines, often match descriptors in order to produce a graph of corresponding features in multiple views. All subsequent stages of such a pipeline strongly depend on the quantity and the quality of these correspondences. In order to give practical insights, the authors evaluate the impact of the choice of descriptor at four stages of such a pipeline: feature matching, geometric verification, image retrieval, and sparse and dense modelling.
They consider RSIFT as a baseline descriptor, along with two advanced variants of it, RSIFT-PCA~\citep{BursucICMR15Kernel} and DSP-SIFT~\citep{DongCVPR15DomainSize}, and they compare these with ConvOpt, DeepDesc, TFeat and LIFT (see Section~4.1). Except for
LIFT, which is an end-to-end detector and descriptor network, SIFT's difference-of-Gaussians (DoG) keypoint detector was used for all descriptors and the learning-based approaches were trained on DoG keypoints.\\

Table~\ref{tab:kprop} shows basic properties of the evaluated descriptors from \citet{SchonbergerCVPR17Comparative}.
We can see that the extraction times of the descriptors vary by an order of magnitude with learned descriptors being up to an order of magnitude slower than handcrafted descriptors, even though the latter run on a GPU.
Among the learning-based features, LIFT by far the slowest and as such it is clearly not a practical alternative for processing millions of images as required by some image-based reconstruction use cases.
Meanwhile matching times depend largely on the size of the descriptor.
Since this size varies by less than a factor of two over the set of descriptors evaluated,
the matching times are also not highly variable.\\

\begin{table}[t]
 \caption{Basic properties of the descriptors evaluated in \citet{SchonbergerCVPR17Comparative}. Per image average timings were measured on the Oxford5k dataset. Extraction time includes detection time. (By courtesy of Johannes~L.~Sch\"{o}nberger.)}
 \label{tab:kprop}
\begin{center}
\bt{l@{\hskip 2mm}r@{\hskip 2mm}r@{\hskip 2mm}r@{\hskip 2mm}r@{\hskip 2mm}r@{\hskip 2mm}r@{\hskip 2mm}r}
\hline
 & RSIFT & RSIFT-PCA & DSP-SIFT & ConvOpt & DeepDesc & TFeat & LIFT \\
\hline
\textit{Dimensionality} & 128 & 80 & 128 &  73 & 128 & 128 & 128 \\
\textit{Size (bytes)} &  128 & 320 & 512 & 292 & 512 &  512 & 512\\
\textit{Platform} &   CPU & CPU & CPU & GPU &  GPU & GPU & GPU \\
\textit{Extraction (s)} & 9.3 & 10.5 & 23.7 & 49.9 & 24.3 & 11.8 & 212.3  \\
\textit{Matching (s)} &  0.14 & 0.11 &  0.14 & 0.10 & 0.14 & 0.14 & 0.14 \\
\hline
\et
\end{center}
\end{table}

Concerning the evaluation of the impact of the choice of descriptor at different stages of the image-based reconstruction pipeline, we summarise the findings of \citet{SchonbergerCVPR17Comparative} as follows.
In agreement with \citet{HeinlyECCV12Comparative}, the paper shows that for all stages of the pipeline, blur, day-night, and large viewpoint changes seriously challenge all descriptors.
The learned descriptors typically outperformed RSIFT in terms of recall, while RSIFT performed better in terms of precision.
Both RSIFT-PCA and DSP-SIFT outperformed the learned features for almost all metrics and matching scenarios tested.
Among the learned descriptors, ConvOpt was found to produce the best overall results and had the lowest variance across the different datasets.\\

To evaluate  the completeness and accuracy of the reconstruction
results, the metrics used in \citet{SchonbergerCVPR17Comparative} were
the number of registered images, the number of 3D points in the sparse SfM map output by COLMAP \citep{SchonbergerCVPR16Structure}, the number of verified image projections of sparse points and their track lengths, the overall reprojection error, the pose accuracy of the camera locations, as well as the number of reconstructed
dense points after multi-view stereo (MVS) reconstruction using CMVS~\citep{YasutakaPAMI10Accurate}.\\

The experiments on several datasets, yielded the following observations: on small or easy datasets, the learned descriptors generally perform on a par with or better than RSIFT in terms of the number of sparse points, the number of image observations, and the mean track length; but they performed worse than RSIFT-PCA and DSP-SIFT on these metrics. However, in terms of the number of registered images and the final dense modelling performance and accuracy metrics, all methods produce roughly the same reconstruction quality.\\

On larger and more challenging datasets, more variation was found when ranking the features using different metrics and datasets.
In spite of the superiority of the learned descriptors over RSIFT observed in raw matching evaluation,
in the reconstruction evaluation RSIFT sometimes performed better and sometimes worse than the learned descriptors.
DSP-SIFT performed the best among all the methods, both in terms of sparse and dense reconstruction results. 
It consistently produced the most complete sparse reconstructions in terms of the number of registered images and reconstructed sparse
points, and its dense models had the most points as a result of accurate camera registration. DSP-SIFT has a slightly higher reprojection error than other methods based on the DoG keypoint detector. This is potentially caused by the descriptor pooling across multiple scales, which improves robustness but also results in less accurate keypoint localisation. LIFT has the largest reprojection error and relatively short tracks on all datasets, indicating inferior keypoint localisation performance as compared to the handcrafted DoG method.\\

Concerning camera-pose estimation, the ground truth was only available for three datasets: two small ones, namely
Fountain and Herzjesu~ \citep{StrechaCVPR08Benchmarking}; and the Quad6K from the Cornell BigSfM dataset \citep{CrandallPAMI13SfM}. On the small datasets all methods performed similarly. On Quad6K, RSIFT performed best, followed by TFeat and DCP-SIFT, with RSIFT-PCA performing worst.
In summary, even if learning approaches advanced, handcrafted features still perform on par or better than recent learned features in the practical context of image-based reconstruction.  \\

\section{Conclusions}
\label{sec:conclusion}

This paper gave an overview of keypoint detectors and descriptors, focussing on local features that are designed to be localised accurately and consistently over time, rather than local features designed for extracting semantics. We traced the evolution of such detectors and descriptors, from classic handcrafted methods through to more recent learning and deep-learning-based methods. Then we discussed existing benchmark papers, which compare most of those methods. While this discussion may be useful for a reader interested in selecting the best off-the-shelf local features type for a given task and data source at the time of writing, it also suggests that better results may be possible by combining ideas from different state-of-the-art methods. We highlight the following findings.
\begin{enumerate}
\item Learned methods are a good choice when image content matters, especially for applications like image matching. 
\item Post-processing descriptors by whitening, power-law normalisation, and $L_2$-normalisation often improves matching results.
\item Among the non-deep-learning-based models, ConvOpt shows good performance  across  different datasets and tasks, but it was outperformed by several recent deep models. The deep models TFeat, L2-Net and HardNet perform well, but they can be improved by (kernel) subspace pooling (SP, KSP) or bilinear pooling, as well as by adding a global loss (TGLoss) or global orthogonal regularization (GOR). 
\item The highest mean average precision on patch verification, matching, and retrieval tasks was attained by DOAP, which directly optimises the average precision instead of using a pairwise or triplet loss.
\item  TConvDet was shown to provide the best keypoint repeatability compared to other detectors and combined with SIFT descriptor provides good matching performance. 
\item  LIFT and SuperPoint are among the few networks that learn keypoint detection,  geometric transformation and keypoint description in a single network trained end-to-end. 
\item  SIFT and many of its variants show high robustness to viewpoint changes, and they remain a good option for most applications.
\item   In addition to the advantages of low memory footprint and matching time, deep-learned binary features, such as binary DOAP, provide competitive results on recent benchmarks.
\item  Extraction of handcrafted descriptors on a CPU is often much faster than extraction of learned descriptors on a GPU, but there are some exceptions. 
\item  Even though learning approaches have advanced to the extent that they now attain the highest mean average precision on matching, recent benchmarks targeting their application in image-based reconstruction and localisation pipelines suggest that handcrafted features still perform just as well or even better than recent deep-learned features on such tasks. However, such benchmarks were conducted before methods like SP, KSP and DOAP were published. 
\end{enumerate}   

Since the applications of computer vision are diverse and the associated data is to a large extent unpredictable, we would argue that learning detectors and descriptors is preferable to manually designing them.
By taking such a learning approach, future computer vision algorithms will best cope with different tasks, imaging modalities and environments.


\bibliographystyle{plainnat}



\end{document}